\documentclass[10pt,twocolumn,letterpaper]{article}

\usepackage[pagenumbers]{cvpr} %
\usepackage{comment}

\usepackage{amsmath,amsfonts,bm}

\def\eqref#1{equation~\ref{#1}}
\def\Eqref#1{Equation~\ref{#1}}

\def\1{\bm{1}}

\def\vb{{\bm{b}}}
\def\vc{{\bm{c}}}

\def\ve{{\bm{e}}}

\def\vg{{\bm{g}}}

\def\vp{{\bm{p}}}

\def\vt{{\bm{t}}}
\def\vu{{\bm{u}}}

\def\vz{{\bm{z}}}

\def\mC{{\bm{C}}}

\def\mE{{\bm{E}}}

\def\mG{{\bm{G}}}

\def\mI{{\bm{I}}}

\def\mY{{\bm{Y}}}
\def\mZ{{\bm{Z}}}

\DeclareMathAlphabet{\mathsfit}{\encodingdefault}{\sfdefault}{m}{sl}
\SetMathAlphabet{\mathsfit}{bold}{\encodingdefault}{\sfdefault}{bx}{n}

\def\gC{{\mathcal{C}}}

\def\sF{{\mathbb{F}}}

\def\sQ{{\mathbb{Q}}}

\def\sS{{\mathbb{S}}}

\newcommand{\E}{\mathbb{E}}
\newcommand{\Ls}{\mathcal{L}}
\newcommand{\R}{\mathbb{R}}

\DeclareMathOperator*{\argmin}{arg\,min}

\DeclareMathOperator{\sign}{sign}

\newcommand{\Z}{\mathbb{Z}}

\def\cE{{\mathcal{E}}}
\def\cG{{\mathcal{G}}}
\def\cQ{{\mathcal{Q}}}

\DeclareMathAlphabet{\mathbbold}{U}{bbold}{m}{n}

\newcommand{\myparagraph}[1]{\vspace{2pt}\noindent{\bf #1}}
\newcommand{\std}[1]{{{$\pm${#1}}}}
\def\LSQ{$\mathbbold{\Lambda}_{24}$-SQ\xspace}
\def\LSQs{$\mathbbold{\Lambda}_{24}$-SQ}
\def\rvphi{\Phi}

\usepackage{xspace}
\usepackage{booktabs}
\usepackage{pdflscape}
\usepackage{multirow}
\usepackage{graphicx}
\usepackage{caption}

\usepackage{array}
\newcolumntype{H}{>{\setbox0=\hbox\bgroup}c<{\egroup}@{}}
\newcommand{\tablestyle}[2]{\setlength{\tabcolsep}{#1}\renewcommand{\arraystretch}{#2}\centering\small}

\usepackage{tikz}
\usetikzlibrary{shapes,backgrounds,calc}

\usepackage{mathtools}

\definecolor{lightgray}{HTML}{e9e9e9}
\definecolor{botticelli}{HTML}{cee1ec}
\definecolor{wepeep}{HTML}{f4d1d5}
\definecolor{sidecar}{HTML}{f3e2b7}
\definecolor{oysterpink}{HTML}{eaccd1}
\definecolor{maize}{HTML}{f7cab1}

\usepackage[rightcaption,wide]{sidecap}
\sidecaptionvpos{figure}{t}
\sidecaptionvpos{table}{t}

\usepackage{colortbl}

\definecolor{cvprblue}{rgb}{0.21,0.49,0.74}
\usepackage[pagebackref,breaklinks,colorlinks,allcolors=cvprblue]{hyperref}

\def\paperID{****} %
\def\confName{CVPR}
\def\confYear{2026}

\title{Spherical Leech Quantization for Visual Tokenization and Generation}

\author{
Yue Zhao$^{1,2}$ \hspace{0.1em}
Hanwen Jiang$^{1,3}$ \hspace{0.1em}
Zhenlin Xu$^{4,}$\footnotemark[2] \hspace{0.2em}
Chutong Yang$^{1}$ \hspace{0.1em}
Ehsan Adeli$^{2,}$\footnotemark[1] \hspace{0.2em}
Philipp Kr\"ahenb\"uhl$^{1,}$\footnotemark[1] \\
$^1$ UT Austin \quad
$^2$ Stanford University \quad
$^3$ Adobe Research \quad
$^4$ Mistral AI \\
{\tt \url{http://cs.stanford.edu/~yzz/npq/}}
}

\begin{document}

\twocolumn[{
\renewcommand\twocolumn[1][]{#1}
\maketitle
\begin{center}
    \vspace{-15pt}
	\centering
    \begin{minipage}[c]{0.32\linewidth}
        \resizebox{\textwidth}{!}{
        
\begin{tikzpicture}[scale=1]
        \begin{scope}[blend group=screen]
        \draw[fill=botticelli,opacity=0.8] (-1,0) ellipse (3cm and 2.4cm);
        \end{scope}
        \begin{scope}[blend group=screen]
        \draw[fill=wepeep,opacity=0.8] (1,0) ellipse (3cm and 2.4cm);
        \end{scope}
        \draw[fill=lightgray] (0,-0.5) ellipse (1.2cm and 1.5cm);
        \draw[fill=sidecar] (0,-0.6) circle (0.8cm);
        \draw[fill=maize] (0.1,-0.7) circle (0.4cm);
        \node (vq) at (3, 1.2) [circle,fill,inner sep=1.2pt]{};
        \node (vq_caption) at (3.0, 0.55) [text width=2cm]{Learned VQ ~\cite{van2017vqvae,yu2022vitvqgan}};
        \node (lfq) at (-0.5, -0.5) [circle,fill,inner sep=1.2pt]{};
        \node (lfq_caption) at (-2.0, -3) {LFQ~\cite{yu2023magvit2}};
        \draw[->] (lfq_caption.north) -- (lfq);
        \node (bsq) at (-0.5, -1) [circle,fill,inner sep=1.2pt]{};
        \node (bsq_caption) at (-0.5, -3) {BSQ~\cite{zhao2025bsq}};
        \draw[->] (bsq_caption.north) -- (bsq);
        \node (fsq) at (-0.4, 0.7) [circle,fill,inner sep=1.2pt]{};
        \node (fsq_caption) at (-3.5, -3) {FSQ~\cite{mentzer2023fsq}};
        \draw[->] (fsq_caption.north) -- (fsq);
        \node (random) at (1.5, -0.2) [circle,fill,inner sep=1.2pt]{};
        \node (random_caption) at (2.3, -3) {Random projection~\cite{chiu2022vqrp}};
        \draw[->] (random_caption.north) -- (random);
        \node (lattice_code) at (-3, 3.4) {\colorbox{botticelli}{Lattice code}};
        \draw[dashed,->] (lattice_code.south) -- (-2.5, 2.078);
        \node (vq) at (2.6, 3.6) {\colorbox{wepeep}{Vector quantization}};
        \draw[dashed,->] (vq.south) -- (2.5, 2.078);
        \node (nonparametric_caption) at (-1.4, 4.0) {\colorbox{lightgray}{Non-parametric quantization}~[\cref{sec:method:lattice}]};
        \draw[dashed,->] (nonparametric_caption.south) -- (-0.5, 0.864);
        \node (shperical_caption) at (0.8, 3.0) {\colorbox{sidecar}{Spherical lattices}~\cite{ericson2001codes}};
        \draw[dashed,->] (shperical_caption.south) -- (0.5, 0.024);
        \node (packing_caption) at (0.6, -4) {\colorbox{maize}{Densest sphere packing lattices}~[\cref{sec:method:spherical}]};
        \draw[dashed,->] (packing_caption.north) -- (0.3, -1.046);
        \node (slq) at (0.1, -0.8) [star,fill,inner sep=1.2pt]{};
        \node (leech_caption) at (0.36, -3.5) {Spherical Leech lattice~[\cref{sec:method:slq}]};
        \draw[->] (leech_caption.north) -- (0.1, -0.9);

      \def\x{1.5}\def\y{1.5}
      \def\e{0.7071}
      \def\sqrt3{1.73205080757}
  \begin{scope}[shift={(-2.5cm, -8.0cm)}]
      \draw[->] (-\x-0.8,0) -- (\x+0.8,0) node[below right] {$x$};
      \draw[->] (0,-\y-0.8) -- (0,\y+0.8) node[above] {$y$};
    
      \draw[orange,fill=sidecar,fill opacity=0.8] (0,0) circle [x radius=1, y radius=1];
    
      \coordinate[circle, draw, inner sep=1pt] (c1) at (\e, \e);
      \node (c1l) at (\e, \e) [above right] {$\vc_1$};
      \coordinate[circle, draw, inner sep=1pt] (c2) at (-\e, \e);
      \node (c2l) at (-\e, \e) [above left] {$\vc_2$};
      \coordinate[circle, draw, inner sep=1pt] (c3) at (-\e, -\e);
      \node (c3l) at (-\e, -\e) [below left] {$\vc_3$};
      \coordinate[circle, draw, inner sep=1pt] (c4) at (\e, -\e);
      \node (c4l) at (\e, -\e) [below right] {$\vc_4$};
      \foreach \dx/\dy in { (0,0), (\e, \e), (-\e, \e),  (-\e, -\e), (\e, -\e), (2*\e,0), (0,2*\e),(-2*\e,0), (0,-2*\e) }
      {
          \begin{scope}[shift={(\dx,\dy)}]
          \draw[dashed] (0,\e)--(\e,0)--(0,-\e)--(-\e,0)--(0,\e);
          \end{scope}
      }
      \node (caption) at (0, -2.7) {BSQ ($d=2$)};
    \node (caption) at (0, -3.2) {simple square lattice};
  \end{scope}

  \begin{scope}[shift={(2.5cm, -8.0cm)}]
      \draw[->] (-\x-0.8,0) -- (\x+0.8,0) node[right] {$x$};
      \draw[->] (0,-\y-0.8) -- (0,\y+0.8) node[above] {$y$};
    
      \draw[orange,fill=sidecar,fill opacity=0.8] (0,0) circle [x radius=1, y radius=1];
    
      \coordinate[circle, draw, inner sep=1pt] (c1) at (0.866, 0.5);
      \node (c1l) at (0.866, 0.5) [above right] {$\vc_1$};
      \coordinate[circle, draw, inner sep=1pt] (c2) at (0, 1.0);
      \node (c2l) at (0, 1.0) [above right] {$\vc_2$};
      \coordinate[circle, draw, inner sep=1pt] (c3) at (-0.866, 0.5);
      \node (c3l) at (-0.866, 0.5) [above left] {$\vc_3$};
      \coordinate[circle, draw, inner sep=1pt] (c4) at (-0.866, -0.5);
      \node (c4l) at (-0.866, -0.5) [above left] {$\vc_4$};
      \coordinate[circle, draw, inner sep=1pt] (c5) at (0, -1.0);
      \node (c5l) at (0, -1.0) [below left] {$\vc_5$};
      \coordinate[circle, draw, inner sep=1pt] (c6) at (0.866, -0.5);
      \node (c6l) at (0.866, -0.5) [below right] {$\vc_6$};
      \foreach \dx/\dy in { (0,0), (0.866, 0.5), (0, 1), (-0.866, 0.5), (-0.866, -0.5), (0, -1.0), (0.866, -0.5) }
      {
          \begin{scope}[shift={(\dx,\dy)}]
          \draw[dashed] (1/\sqrt3,0)--(0.5/\sqrt3,0.5)--(-0.5/\sqrt3,0.5)--(-1/\sqrt3,0)--(-0.5/\sqrt3,-0.5)--(0.5/\sqrt3,-0.5)--(1/\sqrt3,0);
          \end{scope}
      }
    \node (caption) at (0, -2.7) {$\mathbb{A}_2$ (Densest packing at $d=2$)};
    \node (caption) at (0, -3.2) {hexagonal lattice};
  \end{scope}

    \end{tikzpicture}

    }
    \end{minipage}
    \hfill
    \begin{minipage}[c]{0.32\linewidth}
        \includegraphics[width=\linewidth,height=1.5\linewidth]{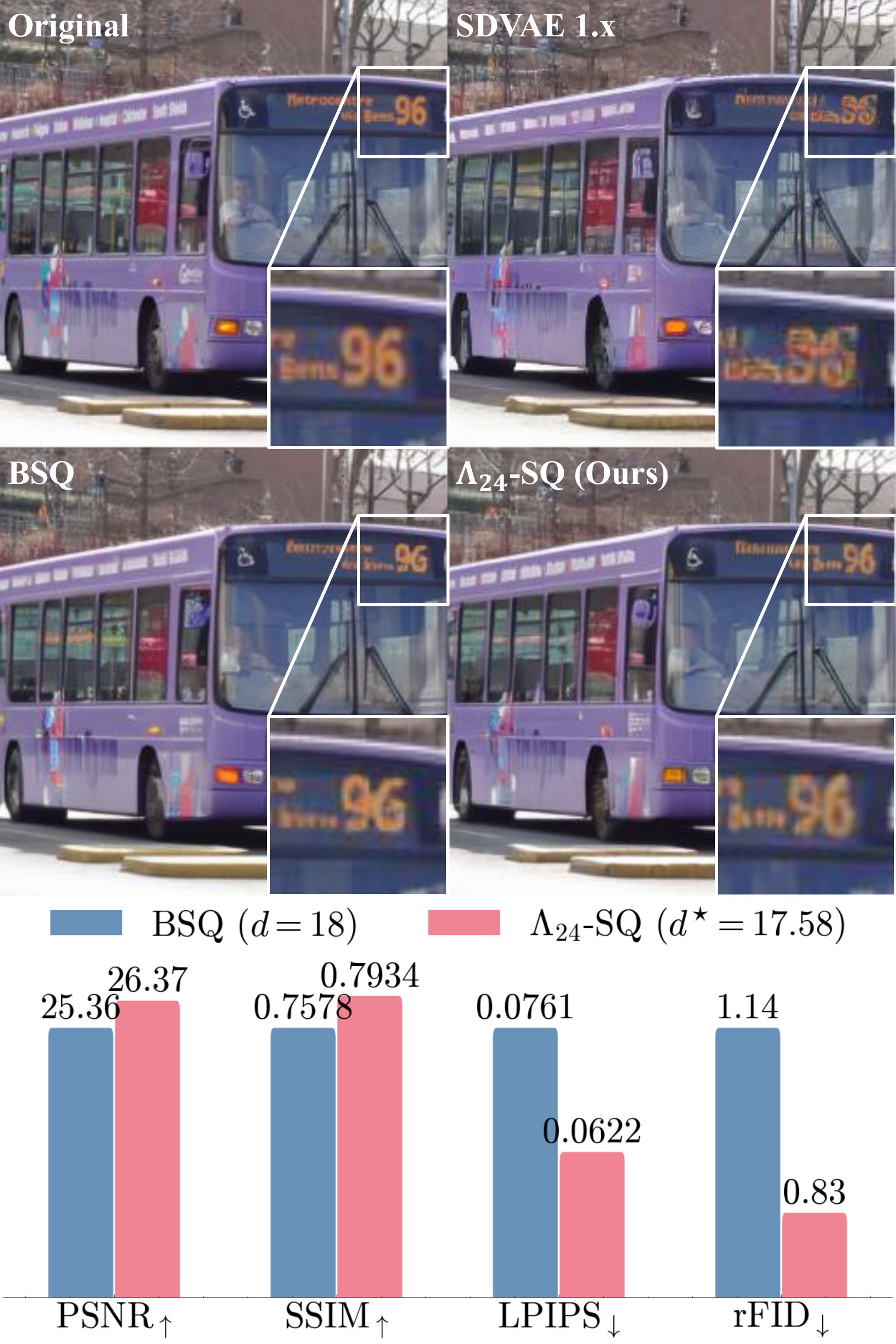}
    \end{minipage}
    \hfill
    \begin{minipage}[c]{0.32\linewidth}
        \includegraphics[width=\linewidth,height=1.5\linewidth]{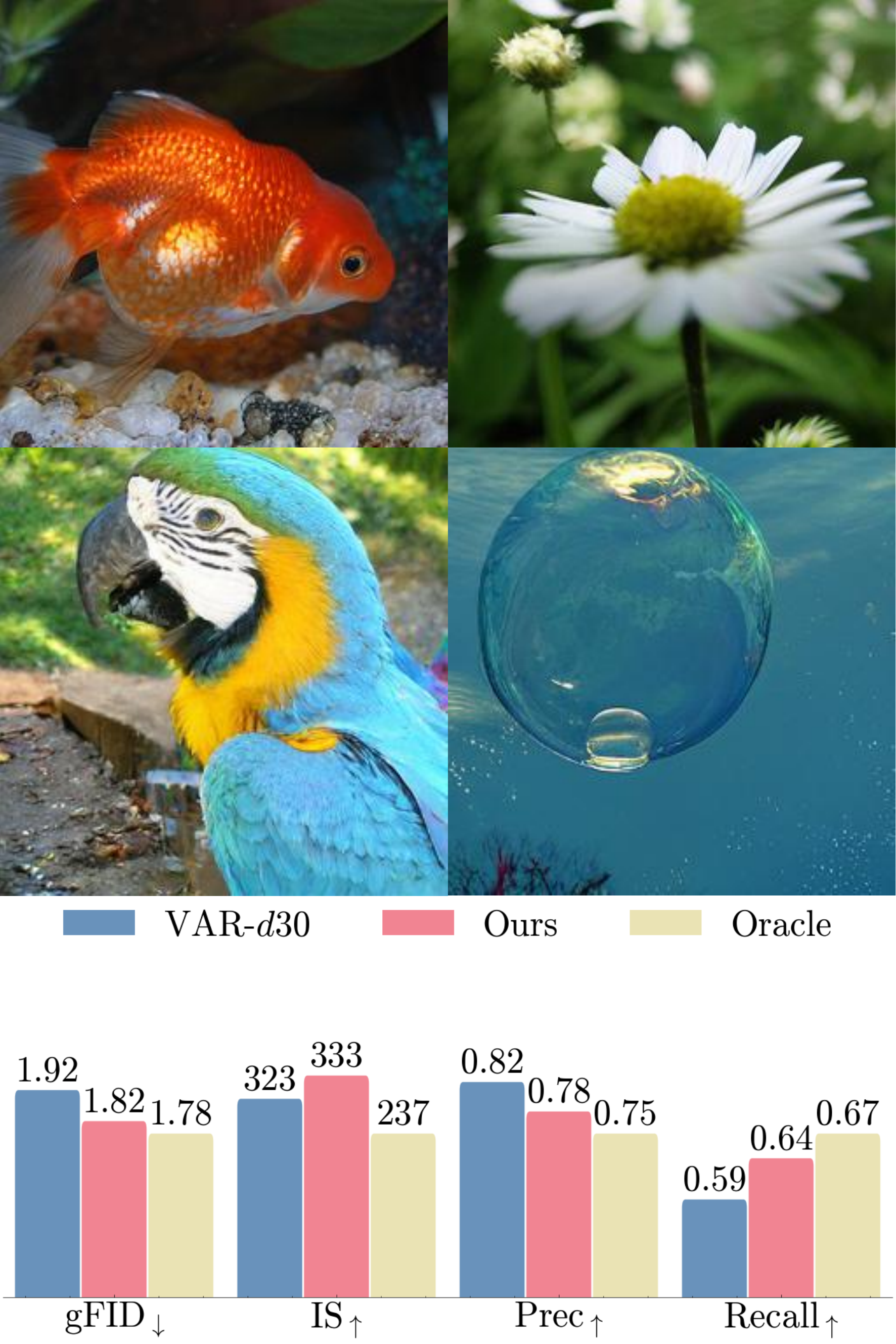}
    \end{minipage}
    \vspace{-10pt}
	\captionof{figure}{
    \small{\textbf{Upper left: A Venn Diagram that contains all definitions and quantization methods covered in this paper}.}
    We provide a unified formulation of various non-parametric quantization methods~\cite{yu2023magvit2,mentzer2023fsq,zhao2025bsq} from a lattice-coding perspective in \Cref{sec:method:lattice}.
    The geometric interpretation of the entropy penalties in \Cref{sec:method:entropy} then leads to a family of densest hypersphere packing lattices (\Cref{sec:method:spherical}).
    Based on the spherical Leech lattice, a $24$-$d$ case of the densest hypersphere packing lattices, we instantiate Spherical Leech Quantization (\LSQ) in~\Cref{sec:method:slq} and apply it to modern discrete auto-encoders (middle) and visual autoregressive models (right). 
    \textbf{Lower left: An illustrative 2D comparison between BSQ and spherical densest-packing lattice quantization ($\mathbb{A}_2$).}
    \textbf{Middle: An auto-encoder with \LSQ outperforms BSQ in image reconstruction and compression (Qualitative results on the top).}
    \textbf{Right: Qualitative and quantitative results of a visual autoregressive generation model with \LSQ on ImageNet-1k.}
    For the first time, we train a discrete visual autoregressive generation model with a codebook of $196,560$ \textit{without bells and whistles} and achieve \textit{an oracle-like performance}.
    }
	\label{fig:overview}
\end{center}
}]

\renewcommand*{\thefootnote}{\fnsymbol{footnote}}
\footnotetext[1]{Equal advising.}
\footnotetext[2]{Work done before joining Mistral AI.}
\renewcommand*{\thefootnote}{\arabic{footnote}}
\begin{abstract}
Non-parametric quantization has received much attention due to its efficiency on parameters and scalability to a large codebook.
In this paper, we present a unified formulation of different non-parametric quantization methods through the lens of lattice coding.
The geometry of lattice codes explains the necessity of auxiliary loss terms when training auto-encoders with certain existing lookup-free quantization variants such as BSQ.
As a step forward, we explore a few possible candidates, including random lattices, generalized Fibonacci lattices, and densest sphere packing lattices.
Among all, we find the Leech lattice-based quantization method, which is dubbed as Spherical Leech Quantization (\LSQ), leads to both a simplified training recipe and an improved reconstruction-compression tradeoff thanks to its high symmetry and even distribution on the hypersphere.
In image tokenization and compression tasks, this quantization approach achieves better reconstruction quality across all metrics than BSQ, the best prior art, while consuming slightly fewer bits.
The improvement also extends to state-of-the-art auto-regressive image generation frameworks.
\end{abstract}
    
\section{Introduction}
\label{sec:intro}

Learning discrete visual tokenization is fundamental to visual compression~\cite{daede2016daala,gersho2012vqsc}, generation~\cite{esser2021vqgan,chang2022maskgit,tian2024var}, and understanding~\cite{bao2022beit}.
Although discrete token-based visual modeling~\cite{bai2024lvm,tian2024var} follows a recipe similar to that of language modeling, we observe a paradox:
Visual information carries much more data than language in quantity and diversity\footnote{Human language conveys information at several tens of bits/s~\cite{shannon1951prediction,piantadosi2011word} while the brain receives visual input at $10^6$ bits/s~\cite{koch2006much}.};
However, the visual vocabulary size of vision models lags far behind that of Large Language Models (LLM)\footnote{The tokenizer has 199,997 elements in GPT-4o~\cite{hurst2024gpt4o} while 129,280 in Deepseek-R1~\cite{guo2025deepseekr1}. Meanwhile, a typical visual codebook is around $1,024\sim16,384$.}.
To bridge this gap, non-parametric quantization (NPQ) methods~\cite{yu2023magvit2,mentzer2023fsq,zhao2025bsq} have recently been proposed, demonstrating codebook scalability and parameter efficiency compared to vector quantization~\cite{gray1984vq,van2017vqvae}.
However, existing NPQ methods have their own flaws and require ad hoc tweaks (\eg, regularization terms), which eventually boil down to the fact that most methods are heuristic and lack a principled design.

In this paper, we propose a simple and effective quantization method, called Spherical Leech Quantization (\LSQ), which scales to a codebook of $\sim200K$ while keeping the training of both visual tokenizers \textit{and} visual autoregressive models as simple as possible. 
\LSQ is theoretically grounded in the intersection of vector quantization and lattice codes. 
We first provide a unified formulation of existing non-parametric quantization methods from the perspective of lattice coding and reinterpret entropy penalties as a lattice relocation problem.
This motivates a family of densest hypersphere packing lattices, among which the Leech lattice in the first shell~\cite{leech1967notes} instantiates the codebook of \LSQ.

Spherical Leech Quantization features the following advantages.
\textbf{(i) Simplicity}:
Thanks to the densest sphere packing principle, \LSQ enables the autoencoder to train with the simplest loss trio (\ie $\ell_1$, GAN, and LPIPS) \textit{without any} regularization terms such as commitment loss and entropy penalties.
\textbf{(ii) Efficiency}: Because of the fixed lattice vectors, \LSQ is excluded from gradient updates, being both memory and runtime efficient.
\textbf{(iii) Effectiveness}. \LSQ effectively pushes the rate-distortion tradeoff frontier.
Specifically, \LSQ-based autoencoder improves rFID from 1.14 to 0.83 compared to BSQ with a slightly smaller effective bitrate ($d^\star=17.58$ \vs $d=18$).
See \Cref{fig:overview}.

Based on \LSQ, we introduce improved techniques in training autoregressive visual generation models with a very large codebook.
For the first time, we train a discrete visual autoregressive generation model with a codebook of $\sim200K$, comparable to frontier language models, \textit{without bells and whistles} (index subgrouping~\cite{yu2023magvit2,zhao2025bsq}, multihead prediction~\cite{han2025infinity}, bit flipping/self-correction~\cite{weber2024maskbit}, \etc.) and achieve a generation FID of 1.82 FID, close to the validation oracle (1.78 FID) on ImageNet-1k.

\section{Preliminaries}
\label{sec:prelim}

\subsection{Visual tokenization and quantization}
\label{sec:prelim:tokenization}

\myparagraph{Visual tokenization.}
Visual tokenization transforms visual input into a set of discrete representations using an auto-encoder architecture and a bottleneck module based on vector quantization (VQ)~\cite{gray1984vq,van2017vqvae}.
In this paper, we consider single images as input for simplicity. 
Given an image $\mI \in \R^{H\times W \times 3}$ and an encoder (decoder) denoted by $\cE$ ($\cG$), we have 
\begin{align}
    & \mI \xrightarrow[\text{encoder}]{\cE(\cdot)} \mZ\in \R^{\left(\frac{H}{p}\times\frac{W}{p}\right)\times d} \xrightarrow[\text{quantizer}]{\cQ_\mathit{VQ}(\cdot)} \hat{\mZ} \xrightarrow[\text{decoder}]{\cG(\cdot)} \hat{\mI},
\end{align}
where $p$ is the spatial downsample factor and the quantizer $\cQ$ assigns each $\vz \in \mZ$ to the closest entry $k^{\star}$ in a learnable codebook $ \mC = [\vc_1, \cdots, \vc_K]\in \R^{K\times d}$,~\ie $ \hat{\vz} = \vc_{k^{\star}} = \argmin_{\vc_k} \| \vz - \vc_{k} \|^2 $.
The entire model $(\cE$, $\cG$, and $\cQ)$ is end-to-end trainable using approximation methods such as straight-through estimator (STE)~\cite{bengio2013ste}, Gumbel Softmax~\cite{jang2017gumbelsoftmax}, or more recent tricks like Rotation Trick~\cite{fifty2025rotation}.
One of the key challenges is to learn the codebook effectively, especially when the codebook size $K$ increases.

\myparagraph{Implicit codebooks.}
Yu \etal introduced a \emph{fixed} implicit codebook $\mC_\mathit{LFQ}=\{\pm1\}^d$~\cite{yu2023magvit2}, where
its best quantizer is binary quantization $\cQ_\mathit{LFQ}(\vz) = \sign(\vz)$.
Binary Spherical Quantization (BSQ)~\cite{zhao2025bsq} further projects the hypercube-shaped codebook onto a unit sphere,~\ie $\mC_\mathit{BSQ}=\{\pm\frac{1}{\sqrt{d}}\}^d$.
Finite Scalar Quantization (FSQ)~\cite{mentzer2023fsq} extends the codebook to multiple values per dimension~\ie $\prod_{i=1}^d\{0,\cdots,\pm \lfloor\frac{L_i}{2}\rfloor\}$.
We refer to these implicit codebook-based methods as \textit{non-parametric quantization (NPQ)}.
Both LFQ and BSQ require an entropy regularization term~\cite{jansen2020coincidence} to encourage high code utilization:
\begin{align}
    \Ls_\mathrm{entropy} = \E \left[ H[q(\vz)] \right] - \gamma H\left[ \E \left[ q(\vz) \right] \right],
    \label{eq:entropy_regularization}
\end{align}
where $q(\vz)\approx \hat{q}(\vc | \vz) = \frac{\exp(-\tau (\vc-\vz)^2)}{\sum_{\vc \in \mC_\mathit{LFQ}} \exp(-\tau (\vc-\vz)^2)}$ is a soft quantization approximation~\cite{agustsson2017soft}.

\begin{table*}[tb]
    \centering
    \caption{\small{\textbf{Comparisons of different non-parametric quantization methods.}} we overload $\prod$ for both scalar product and Cartesian product. }
    \label{tab:lattice_compare}
    \vspace{-10pt}
    \tablestyle{5pt}{1.05}
    \begin{tabular}{cccccc}
        \toprule
         Method & LFQ~\citep{yu2023magvit2}  & FSQ~\citep{mentzer2023fsq}  & BSQ~\citep{zhao2025bsq} & $\sF_d(N)$-SQ & $\mathbbold{\Lambda}_{24}$-SQ (this paper) \\
         \midrule
         Input range & $\R^{d}$ & $\prod_{i=1}^d (-\frac{L_i}{2}, \frac{L_i}{2})$ &  $\sS^{d-1}$ & $\sS^{d-1}$ & $\sS^{23}$ \\
         Code vectors & $\{\pm 1\}^{d}$ & $\prod_{i=1}^d\{-\lfloor\frac{L_i}{2}\rfloor,\cdots,\lfloor\frac{L_i}{2}\rfloor\}$ & $\{\pm \frac{1}{\sqrt{d}}\}^{d}$ & \S\ref{sec:method:spherical} and \S\ref{sec:appendix:fibonacci} & $ \frac{1}{\sqrt{32}} \bigcup_{\{2,3,4\}} \mathbbold{\Lambda}_{24}(2)_s $ \citep{conway2013sphere}  \\
         Codebook size ($|\gC|$) & $2^d$ & $\prod_{i=1}^d L_i$ & $2^d$ & $N$ & 196,560 \\
         Minimum distance ($d_\mathrm{min}$) & 2 & 1 & $\frac{2}{\sqrt{d}}$ & $\delta_\mathrm{min}(N)$ & $\frac{\sqrt{3}}{2}$ \\
         \bottomrule
    \end{tabular}
\end{table*}

Each of the NPQ variants has its own pros and cons.
(1) LFQ is easiest to implement, but the computational cost for entropy increases exponentially.
(2) BSQ provides an efficient approximation with a guaranteed bound, but still suffers from codebook collapse without proper entropy regularization.
(3) FSQ does \emph{not} need such complex regularization, but the way to obtain the number of levels per channel ($L_1,\cdots,L_d$) is somewhat heuristic~\cite{mentzer2023fsq}.
Although the geometric landscape of all these quantization methods varies (Figure 2 in~\cite{zhao2025bsq}), we show in the upcoming chapter that they can be interpreted from the same lattice coding perspective.
This unified formulation further motivates us to develop a novel non-parametric quantization variant that is both theoretically sound and implementation-wise easy.

\subsection{Lattice-based codes}
\label{sec:prelim:lattice}

\myparagraph{Lattice.}
A $d$-dimensional lattice is defined by a discrete set of points in $\R^d$ that constitutes a group.
In particular, the set is translated such that it includes the origin.
Mathematically, an $d$-dimensional lattice $\mathbbold{\Lambda}_n$ is represented by
\begin{align}
    \label{eq:lattice}
    \mathbbold{\Lambda}_d = \{ \bm{\lambda} \in \R^{d} | \bm{\lambda} = \mG \vb \},
    \mG = \begin{bmatrix} | & | &  & | \\ \vg_1 & \vg_2 & \cdots & \vg_d \\  | & | &  & |  \end{bmatrix},
\end{align}
where $\mG \in \R^{d\times d} $ is called the \textit{generator matrix}, comprising of $d$ \textit{generator vectors} $\vg_i$ in columns, and $\vb \in \Z^{d} $.

\myparagraph{Lattice-based codes.}
$\mathbbold{\Lambda}_d$ in \cref{eq:lattice} has infinite elements by definition.
In practice, we include additional \textcolor{blue}{constraints} so that the new set is enumerable:
\begin{align}
    \label{eq:lattice_code}
    \mathbbold{\Lambda}_d = \{ \bm{\lambda} \in \R^{d} | \bm{\lambda} = \mG \vb, \textcolor{blue}{f(\bm{\lambda})=c_1, h(\bm{\lambda})\leq c_2} \}.
\end{align}
Particularly, spherical lattice codes~\cite{ericson2001codes} refers to a family of lattice-based codes with a constant squared norm,~\ie $\mathbbold{\Lambda}_{d,m}=\{ \bm{\lambda} \in \R^{d} | \bm{\lambda} = \mG \vb, \textcolor{blue}{\|\bm{\lambda}\|^2=m} \} $.
We will see more examples in \Cref{sec:method:spherical}.
Besides, we define the quantizer associated with a lattice $\mathbbold{\Lambda}$ by
$ \cQ_{\mathbbold{\Lambda}}(\vz) = \argmin_{\vt \in \mathbbold{\Lambda}} \| \vz - \vt \| $,
which offers a bridge to vector quantization (\Cref{sec:prelim:tokenization}).

\section{Method}
\label{sec:method}

\subsection{Non-parametric quantization as lattice coding}
\label{sec:method:lattice}

From the perspective of the lattice-based codes defined in \cref{eq:lattice_code}, we can describe all variants of non-parametric quantization methods~\cite{yu2023magvit2,mentzer2023fsq,zhao2025bsq} in the same language, despite the varying geometric landscapes. 

\textit{(i) Vanilla Lookup-Free Quantization}~\citep{yu2023magvit2}:
\begin{align}
    \mG &= \begin{bmatrix} | & | &  & | \\ \ve_1 & \ve_2 & \cdots & \ve_d \\  | & | &  & | \end{bmatrix} \triangleq \mI_d, \\
    f_1(\bm{\lambda}) &= \| \bm{\lambda} \|_0 = d,
    f_2(\bm{\lambda}) = \| \bm{\lambda} \|_1 = d.
    \label{eq:lattice_code_constraint_lfq}
\end{align} 
Here, $\ve_i$ is the standard basis vector, taking the value of $1$ at the $i$-th index and $0$ elsewhere. 
The constraints in~\cref{eq:lattice_code_constraint_lfq} are equivalent to saying that $\bm\lambda_i = \pm 1$ for all $i$.

\textit{(ii) Finite Scalar Quantization}~\citep{mentzer2023fsq}:
For simplicity, we consider the special case where any $L_i$ equals $L$.
\begin{align}
    \mG = \mI_d, h(\bm{\lambda}) = \| \bm{\lambda} \|_\infty \leq \frac{L}{2}.
\end{align} 

\textit{(iii) Binary Spherical Quantization}~\citep{zhao2025bsq}:
\begin{align}
    \mG = \frac{1}{\sqrt{d}} \mI_d, 
    f_1(\bm{\lambda}) = \| \bm{\lambda} \|_0 = d,
    f_2(\bm{\lambda}) = \| \bm{\lambda} \|_2 = 1.
   \label{eq:lattice_code_constraint_bsq}
\end{align} 
Although it appears that \cref{eq:lattice_code_constraint_bsq} is simply a scaled version of \cref{eq:lattice_code_constraint_lfq}, it is worth noting that the input range varies: $\vz \in \sS^{d-1}$ in BSQ while $\vz \in \R^{d}$ in LFQ.

\textit{(iv) Random-projection Quantization (RPQ)}~\citep{chiu2022vqrp}.
RPQ initializes the codebook $\{\vp_1, \cdots, \vp_N\}$ using a standard normal distribution, followed by an $\ell_2$ normalization.
Due to the codebook existence, strictly speaking, RPQ does not belong to lookup-free quantization by definition.
Nevertheless, we can still include it in the same picture, where the generator matrix and contraints look like the following:
\begin{align}
    \mG = \begin{bmatrix} | & | &  & | \\ \vp_1 & \vp_2 & \cdots & \vp_N \\  | & | &  & |  \end{bmatrix}, f(\bm{\lambda}) = \| \bm{\lambda} \|_2 = 1,
\end{align} 
where $\mG \in \R^{d\times N}$ slightly abuses the definition, $\vp_i$ follows a projected normal distribution $\vp_i\sim\mathcal{PN}_{d}(0,\mI)$.

\subsection{Entropy regularization as lattice relocation}
\label{sec:method:entropy}

We review the entropy regularization term from the perspective of lattice coding.
We give a geometric interpretation of the entropy regularization and show that the two subterms correspond to pushing the input point towards the lattice points and finding an \textit{optimal} configuration of the lattice. 

\myparagraph{Re-interpretating entropy regularization.}
The \underline{first term} in \cref{eq:entropy_regularization} minimizes the entropy of the distribution that $\vz$ is assigned to one of the codes.
This means that every input should be close to one of the centroids instead of the decision boundary\footnote{This principle is also known as the ``cluster assumption''~\cite{chapelle2005semi,grandvalet2004semi}.}.
This becomes less of an issue since the codebook of interest is huge, 
exemplified by $\gamma$ being greater than 1 as reported in~\cite{jansen2020coincidence,yu2023magvit2}.
Ablation studies in BSQ~\cite{zhao2025bsq} also reveal that we can omit $ \E \left[ H[q(\vz)] \right] $ -- but not $H[\E[q(\vz)]]$ -- while achieving a similar performance.

The \underline{second term} maximizes the entropy of the assignment probabilities averaged over the data, which favors ``class balance''~\cite{krause2010discriminative}.
Assuming that $\vz$ has a uniform distribution over its input range, we have
$\E[q(\vz)] = P(q(\vz)=\vc_k)=\int_{V_k} d\vz = | V_k| $, where $V_k$ is the Voronoi region for the codeword $\vc_k$.
$H[\E[q(\vz)]]$ is maximized when all $V_k$ have equal volumes.

\myparagraph{FSQ implicitly maximizes entropy.}
The interpretation explains why FSQ does not suffer from codebook collapse even without entropy penalties.
Given an input $\vz \in \R^{d}$, FSQ first applies a bounding function $f$, and then rounds to the nearest integers,
\begin{align}
    \vz \xrightarrow[\text{bound}]{f(\cdot)} \bar{\vz} = \lfloor \frac{L}{2} \rfloor \tanh(\vz) \xrightarrow[\text{quantize}]{\cQ_\mathit{FSQ}(\cdot)} \hat{\vz} = \mathrm{round}(\bar{\vz}).
\end{align}
Therefore, the input range is $(-L/2, L/2)$ and all integer points within this range are valid lattice points\footnote{When $L=2,d=3$, this is the well-known ``simple cubic'' or ``primitive cubic'' lattice in crystallography; We will see this again in \Cref{sec:method:spherical}.}.
The codebook size is $L^d$, often in the range of $2^8\sim2^{16}$.
The Voronoi cell for each lattice point $\hat{\vz}$ is a unit-length hypercube $ \prod_{i=1}^d [\hat{z}_i - \frac{1}{2}, \hat{z}_i + \frac{1}{2}) $, implicitly complying with the entropy maximization principle. 

\myparagraph{LFQ requires explicit entropy maximization.}
Although the Voronoi cell for each point in LFQ is also identical, the range of input and quantized output is unbounded.
This breaks the uniform distribution assumption, thus requiring explicit regularization.

\myparagraph{What's left?}
BSQ is missing so far.
Since its input lies on a hypersphere, BSQ has to be treated separately.
We will discuss the lattice relocation problem for spherical lattice codes in \Cref{sec:method:spherical}, where BSQ is one such code.

\subsection{Spherical lattices and hypersphere packing}
\label{sec:method:spherical}

\myparagraph{Spherical lattices.}
The overall pipeline is written as follows:
\begin{align}
    \label{eq:spherical_code}
    \vz\in\R^{d} \xrightarrow[\text{normalize}]{\mathrm{norm}(\cdot)} \tilde{\vz} =\frac{\vz}{\| \vz \|}  \xrightarrow[\text{quantize}]{\cQ_{\mathbbold{\Lambda}}(\cdot)} \hat{\vz} = \cQ_{\mathbbold{\Lambda}}(\tilde{\vz}),
\end{align}
where $\mathrm{norm}(\cdot)$ is another way of ``bounding'', and the Voronoi regions now take arbitrary shapes on the hyperspherical shell.
For simplicity, we study a surrogate problem that approximates the Voronoi regions by placing $N$ $d$-dimensional balls with varied radii\footnote{This will leave some holes, but we believe that the total volume of holes is negligible compared to the balls.}, where $N$ is the cardinality of the lattice in which we are interested.

\myparagraph{Entropy maximization as dispersiveness pursuit.}
The entropy maximization term corresponds to finding the \textit{most dispersive} configuration to relocate the balls.
Sloane \etal formally state this problem of placing $ N $ points on a  $d$-dimensional sphere to maximize the minimum distance (or angle) between any pair of points in \cite{sloane2000spherical}.
This problem generalizes the \textit{Tammes' problem}~\cite{tammes1930origin} in dimensions greater than $3$.
Mathematically, we write this max-min problem as 
$ \max_{\vc_1,\cdots,\vc_{N}\in \sS^{d-1}} \underbrace{\min_{1\leq j < k \leq N} \textit{distance}(\vc_j, \vc_k)}_{\delta_\mathrm{min}(N)} $,
where we denote $\delta_\mathrm{min}(N)$ for future empirical analysis.

\myparagraph{Entropy maximization as hypersphere packing.}
An alternative way is to assume \textit{equal radii} for all hyperspheres and find the \textit{densest sphere packing}~\cite{conway2013sphere}.
The best known results in dimensions 1 to 8, 12, 16, and 24 are summarized in \Cref{tab:dense_packing_lattices}, where $\mathbb{Z}$ to $\mathbb{E}_8$ and $\mathbbold{\Lambda}_{24}$ have been proved optimal among all lattices~\cite{conway2013sphere,cohn2017sphere}.

\begin{table}[!tb]
    \centering
    \caption{
    \small{\textbf{Best known results for dense packing.}
    The content is adapted from Table 1.1 in~\cite{conway2013sphere}.
    }}
    \vspace{-10pt}
    \label{tab:dense_packing_lattices}
    \tablestyle{2pt}{1.05}
    \begin{tabular}{c|ccccccccccc}
        \toprule
        Dimension & 1 & 2 & 3 & 4 & 5 & 6 & 7 & 8 & 12 & 16 & 24 \\
        \midrule 
        densest packing & $\mathbb{Z}$ & $\mathbb{A}_2$ & $\mathbb{A}_3$ & $\mathbb{D}_4$ & $\mathbb{D}_5$ & $\mathbb{E}_6$ &  $\mathbb{E}_7$ & $\mathbb{E}_8$ & $\mathbb{K}_{12}$ & $\mathbbold{\Lambda}_{16}$ & $\mathbbold{\Lambda}_{24}$ \\
        \bottomrule
     \end{tabular}
\end{table}

Given these basics, we now propose a few candidates.

\begin{figure}[!b]
    \vspace{-15pt}
    \centering
    \begin{subfigure}[t]{0.48\linewidth}
      \centering
        \includegraphics[width=\linewidth]{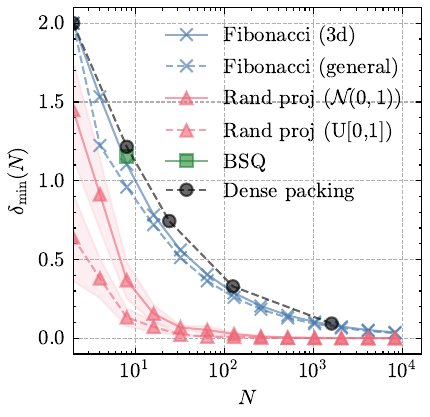}
        \vspace{-15pt}
        \caption{$d=3$.}
        \label{fig:delta_vs_N_3D}
    \end{subfigure}%
    \begin{subfigure}[t]{0.48\linewidth}
      \centering
        \includegraphics[width=\linewidth]{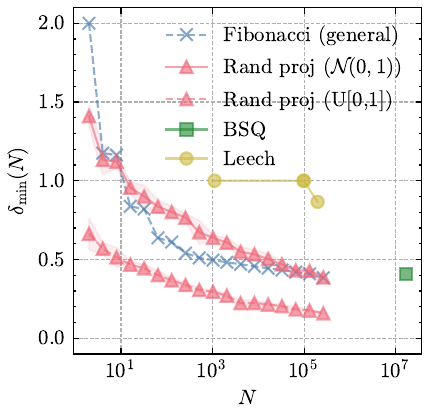}
        \vspace{-15pt}
        \caption{$d=24$.}
        \label{fig:delta_vs_N_24D}
    \end{subfigure}%
    \vspace{-10pt}
    \caption{\small{\textbf{$\delta_\mathrm{min}(|\mathcal{C}|)$ \wrt $|\mathcal{C}|$ across different lattices discussed in \Cref{sec:method:slq} in low dimensions ($d=3$) and high dimensions ($d=24$).}}
    The advantage of the densest sphere packing lattices over other candidates is more visible in higher dimensions.
    }
    \label{fig:delta_vs_N}
\end{figure}

\textit{(i) Random projection lattice}
follows RPQ in \Cref{sec:method:lattice} where the projected normal distribution initializes points.
We use it as a baseline to compare the dispersiveness of different candidate lattice codes in \Cref{fig:delta_vs_N}, which turns out to be surprisingly strong in higher dimensions.

\textit{(ii) Fibonacci lattice} constructs points that \textit{``are evenly distributed with each of them representing almost the same area"}~\citep{gonzalez2010measurement} in a unit square $[0,1)^2$.
We can map this point distribution to a unit-length sphere $\sS^{2}$ using cylindrical equal-area projection.
From Figure~\ref{fig:delta_vs_N_3D}, $\delta_\mathrm{min}(N)$ achieved by this 3D spherical Fibonacci lattice is close to the known densest packing~\cite{sloane2000spherical} and much better than random projection.
We explore its high-dimensional generalization with the hyperspherical coordinate system inspired by~\cite{stackexchange3297830}, denoted by $\sF_d(N)$.
Construction details are left in~\Cref{sec:appendix:fibonacci}.

\textit{(iii) Densest sphere packing lattice} has been introduced and summarized in \Cref{tab:dense_packing_lattices}.
We pay particular attention to the \textit{Leech lattice} $\mathbbold{\Lambda}_{24}$~\cite{leech1967notes}.
$\mathbbold{\Lambda}_{24}$ can be constructed in many ways~\cite{conway1982twenty} and we use the most convient way to calculate.
The vectors in the first shell have a minimal norm $\sqrt{32}$ and fall into three types; we summarize their shapes and numbers in \Cref{tab:leech_lattice_shape} and provide more details in~\Cref{sec:appendix:leech}.
Normalizing these $196,560$ vectors in the first shell to unit length results in the \textit{Spherical Leech Quantization (\LSQ)} codes.
We can easily get $\delta_\mathrm{min}(|\frac{1}{\sqrt{32}}\mathbbold{\Lambda}_{24}(2)_{s}|) = 1$ for $s=2,3,4$ and $\delta_\mathrm{min}(|\frac{1}{\sqrt{32}}\bigcup_{\{2,3,4\}}\mathbbold{\Lambda}_{24}(2)_s|) = \frac{\sqrt{3}}{2}$.
From \Cref{fig:delta_vs_N_24D}, $\delta_\mathrm{min}(N)$ is much larger than all other candidates.

\begin{table}[!tb]
    \centering
    \caption{
    \small{\textbf{Vectors in the first shells of the Leech lattice.} $ \mathbbold{\Lambda}_{24}(n)_i $ indicates the Leech vectors of norm $2n$ (or type $n$ and shape $i$; the signs are suppressed for simplicity. The table is extracted from Table 4.13 in~\cite{conway2013sphere}.
    }}
    \vspace{-10pt}
    \label{tab:leech_lattice_shape}
    \tablestyle{4.2pt}{1.05}
    \begin{tabular}{lll|lll}
    \toprule
    Class     &  Shape  &  Number & Class     &  Shape  &  Number \\
    \midrule
    $\mathbbold{\Lambda}_{24}(0)_1$     &  $(0^{24})$ & 1 & $\mathbbold{\Lambda}_{24}(2)_3$     &  $(3^{1}1^{23})$  & $2^{12}\cdot 24$  \\
    $\mathbbold{\Lambda}_{24}(2)_2$     &  $(2^{8}0^{16})$  & $2^7\cdot 759$ & $\mathbbold{\Lambda}_{24}(2)_4$     &  $(4^{2}0^{22})$  & $2^2 \begin{pmatrix} 24 \\ 2 \end{pmatrix}$  \\
    \bottomrule
    \end{tabular}
\end{table}

\myparagraph{BSQ are not the densest packing lattice codes.}
Before we conclude this chapter, we compare \LSQ with BSQ, the prior art, in Table~\ref{tab:bsq_vs_leech}.
Since the codebook size takes $\log_2(196,560)\approx 17.58$ bits, we use BSQ with $d=18$.
\LSQ increases $\delta_\mathrm{min}(|\mathcal{C}|)$ by more than 80\% ($0.471 \rightarrow 0.866$), indicating its superiority.
Empirical results in Section~\ref{sec:exp} also support that \LSQ enables a simple loss design such that the entropy regularization term can be omitted.  
Comparing \Cref{fig:delta_vs_N_3D,fig:delta_vs_N_24D}, we also conclude that the improvement in $\delta_\mathrm{min}(|\mathcal{C}|)$ of BSQ over the random lattice baseline decreases when $d$ increases.

\subsection{Spherical Leech Quantization in practice}
\label{sec:method:slq}

\myparagraph{Instantiation.}
\LSQ follows the pipeline of \cref{eq:spherical_code} with the lattice being $ \frac{1}{\sqrt{32}} \bigcup_{\{ 2,3,4\}}\mathbbold{\Lambda}_{24}(2)_s $.
Despite the huge codebook size, because the lattice vectors are fixed, we can use tiling and JIT-compiling techniques to reduce both memory and runtime costs compared to vanilla VQ.

\myparagraph{Accomodating smaller codebooks.}
In some cases with less data, a codebook size of $196,560$ may be too large. 
We can also take one type of $\mathbbold{\Lambda}_{24}(2)_s$ or its subset so that the codebook size range $1,104\sim 98,304$.

\begin{table}[!tb]
    \centering
    \caption{\small{\textbf{Comparison between the proposed Spherical Leech Quantization (\LSQ) and BSQ~\citep{zhao2025bsq}.}}}
    \label{tab:bsq_vs_leech}
    \vspace{-10pt}
    \tablestyle{1.8pt}{1.05}
    \begin{tabular}{c|cc}
         \toprule
         Method & BSQ~\citep{zhao2025bsq} & $\mathbbold{\Lambda}_{24}$-SQ (this paper) \\
         \midrule
         Input range & $\sS^{17}$ & $\sS^{23}$ \\
         Code vectors & $\{\pm \frac{1}{\sqrt{18}}\}^{18}$ & See \Cref{tab:lattice_compare}  \\
         Codebook size & $2^{18}=262,144$ & $196,560\approx2^{17.58}$ \\
         $\delta_\mathrm{min}(|\mathcal{C}|)$ & $\frac{2}{\sqrt{18}}\approx0.471$ & $\frac{\sqrt{3}}{2}\approx0.866$ \\
         \midrule
         \multirow{2}{*}{AR output} & \multicolumn{1}{l}{(1) $262,144$-way logits} & \multicolumn{1}{l}{(1) $196,560$-way logits} \\
          & \multicolumn{1}{l}{(2) $18\times$ binary logits}  & \multicolumn{1}{l}{(2) $24\times$ nonary logits}  \\
         Self-correct & bitwise flip & $9$-itwise toggle \\
         \bottomrule
    \end{tabular}
\end{table}

\subsection{Integration with other quantization techniques}
\label{sec:integration}

\myparagraph{Multi-scale residual quantization.}
Since \LSQ is an in-place replacement of VQ, we can use it in combination with other techniques, such as multiscale quantization~\cite{juang1982msvq} and residual quantization~\cite{barnes1996rvq,lee2022rq}.
In this paper, we integrate \LSQ into the VAR tokenizer~\cite{tian2024var} for image generation, allowing for direct comparison with quantization methods such as VQ in VAR~\cite{tian2024var} and BSQ in Infinity~\cite{han2025infinity}.

\myparagraph{Aligning with vision foundation models.}
Better reconstruction does not always lead to better generation quality~\cite{yu2023magvit2,mentzer2023fsq,gupta2024walt,ramanujan2024worse}. 
VAVAE~\cite{van2017vqvae} proposes to address this reconstruction-generation dilemma by aligning latent embeddings with vision foundation models.
We use the VF loss~\cite{yao2025vavae} between the latent embedding \textit{before} quantization and the feature extracted from a pretrained DINOv2~\cite{oquab2023dinov2}.
 
\section{Autoregression with a Very Large Codebook}
\label{sec:generation}

\subsection{Representing the codebook mapping}

\myparagraph{Preliminaries.}
As NPQ scales up the effective size of the codebook, effectively representing the codebook mapping for the autoregressive models becomes a big issue.
The most straightforward way is to represent each code by a unique index.
It uses an embedding matrix $\mE\in\R^{|V|\times D}$ to map each index to a vector and an unembedding matrix $\mE'\in\R^{D \times |V|}$ to get the final logits of dimension $|V|$ for a simple classification problem.
Memory cost and training stability are the two biggest challenges.

\begin{SCfigure*}[1][!tb]
    \centering
    \includegraphics[width=1.6\linewidth]{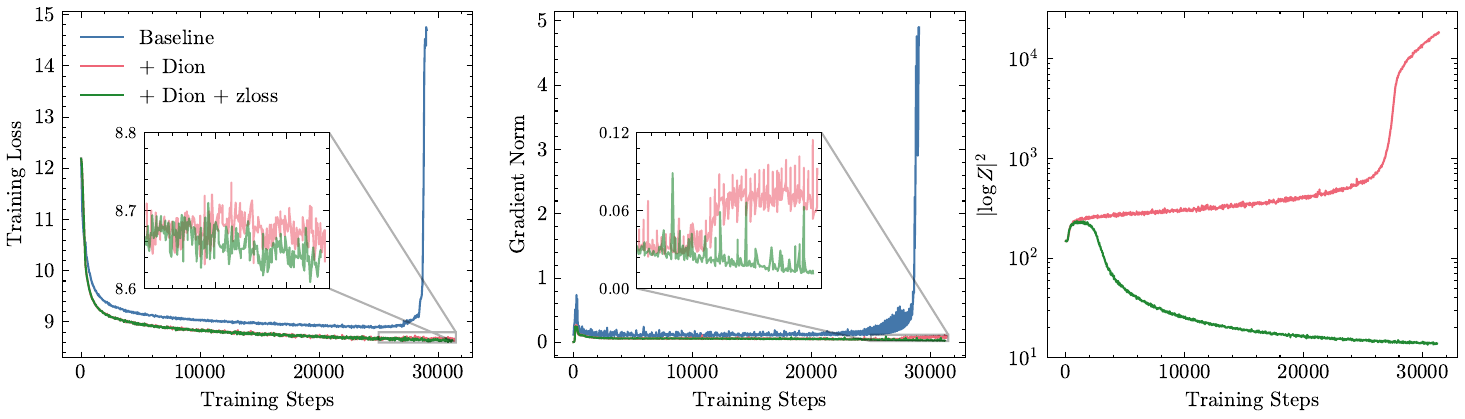}
    \hspace{5pt}
    \caption{
    \small{\textbf{Training curve for a 16-layer $\infty$-CC model.}}
    The Dion optimizer addresses the problem of exploding gradient norm.
    Z-loss effectively regularizes $|\log{Z}|^2$ and smoothens the loss and gradient curve, leading to a lower training loss.
    }
    \label{fig:training_curve_375m}
    \vspace{-15pt}
\end{SCfigure*}

There are two more solutions: (1) index subgroup~\citep{yu2023magvit2} and (2) bitwise operation~\citep{weber2024maskbit,han2025infinity}.
The former is compatible with the autoregression framework, but the resulting sequence length grows linearly \wrt the number of groups.
Han~\etal model bitwise tokens with multiple BCE losses \textit{in parallel} in~\cite{han2025infinity}.
However, it only applies to LFQ/BSQ and relies on bitwise self-correction to mitigate the train-test discrepancy.
In the following, we show our improvements to accommodate the family of spherical lattice codes.

\myparagraph{Simple classification with memory optimization.}
We adopt the cut cross entropy (CCE)~\citep{wijmans2025cce} to address memory consumption.
Since the visual auto-regressive models are trained from scratch, we use Kahan summation~\citep{kahan1965pracniques} to maintain numerical stability, as suggested by~\cite{wijmans2025cce}.
We leave the training techniques in \Cref{sec:ar:train}. 

\myparagraph{Factorized $d$-itwise prediction.}
Densest sphere-packing lattices like $\mathbb{E}_8$ and $\mathbbold{\Lambda}_{24}$ take integer values, but all possible values go beyond binary~\cite{zhao2025bsq}. 
We generalize the concept of bitwise prediction and propose a factorized $d$-it\footnote{Short for \textit{$\log$ base-$d$ unit}, analogous to bit for $\log_2()$ and nat for $\ln()$.} prediction.
Assuming independence across channels, the joint log-probability of one lattice code $\vc^{(1:d)}$ is approximated by the sum of the log-probabilities of each dimension,~\ie
\begin{align}
\log p(\vc^{(1:d)})\approx \sum_{i}^d \log p(\vc^{(i)}),
\label{eq:factorized}
\end{align}
where $p(\vc^{(i)})$ denotes the probability of the $i$-th element of the $d$-dim lattice codes.
For \LSQ, we use $24$ $9$-way classification heads to include all possible values $\{-4,\cdots,4\}$.
\textit{$d$-itwise self-correction} is also possible by toggling any element with a certain probability, though we do not explore it in this paper.

\begin{SCfigure}[1][!tb]
    \centering
    \includegraphics[width=0.5\linewidth]{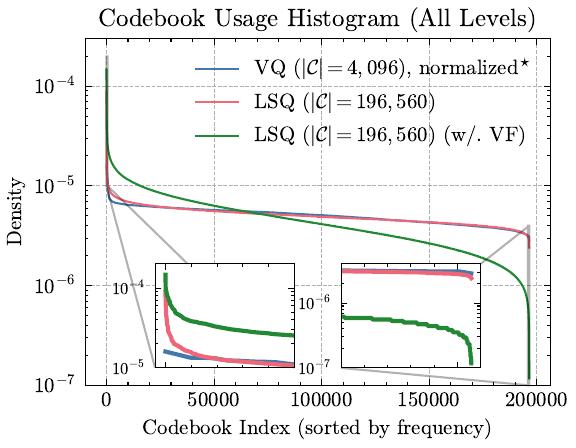}
    \caption{
    \small{\textbf{Codebook usage histogram.}}
    The imbalance in a huge codebook calls for dedicated training tricks in~\S\ref{sec:ar:train}.
    Usage is computed on IN-1k val-50k over 10 VAR levels. 
    $y$-axis in log scale.
    $^\star$: 4,096 VQ codebook indices and density are normalized for illustrative purposes.
    }
    \label{fig:codebook_histogram}
\end{SCfigure}

\subsection{Training}
\label{sec:ar:train}

We train a 16-layer Infinity model but observe a consistent increase in gradient norms and explosion of loss (the blue curve in \Cref{fig:training_curve_375m}\footnote{The loss explosion occurs earlier when the model goes to 1B.}).
A natural hypothesis for this is about the large codebook\footnote{Also, we drop the entropy regularization that promotes class balance.}: We plot the codebook usage on ImageNet-val, sorted by density, in~\Cref{fig:codebook_histogram}.
For the standard VQ codebook, the largest frequency and smallest frequency are within the same order of magnitude ($\frac{7.69e-4}{1.37e-4}\approx5.6$).
For \LSQ's large codebook, the ratio between the most frequent index and the least frequent one surges to $\frac{8.90e-5}{2.41e-6}\approx37$.
The imbalance is more visible after the VF alignment (\cref{sec:integration}).
We further hypothesize that it prevents prior visual auto-regressive models from utilizing a large visual codebook for generation, although the low utilization problem during reconstruction appears to be fixed~\cite{shi2025ibq,zhu2024vqganlc}.

Despite the difficulty, we recognize that this is not a unique issue in visual generation.
An unbalanced, large codebook is common when training large language models~\cite{chowdhery2023palm,wortsman2024small,olmo2025olmo2}.
Therefore, we borrow two simple and effective techniques, namely Z-loss~\citep{chowdhery2023palm} and improved optimization with orthonormalized matrix updates~\cite{jordan2024muon,liu2025muon}.

\myparagraph{Z-loss}~\citep{chowdhery2023palm} prevents the final output logits from exploding.
Namely, $  \Ls_{Z} = \alpha | \log{Z} |^2 =  \alpha \left| \log\left( \sum_{i}^V \exp(z_i) \right) \right|^2$,
where we set $\alpha$ to be $ 10^{-4} $ as in~\cite{olmo2025olmo2}.

\myparagraph{Distributed Orthonormalized Updates}.
We use Dion optimizer~\citep{ahn2025dion} for all weight tensors greater than 1D and Lion~\citep{chen2023lion} for 1D weight tensors and the [un]embedding layers.
The unembedding layer is updated with a learning rate scaled by $1/\sqrt{d_\mathrm{in}}$, where $d_\mathrm{in}$ is its input dimension.

From~\Cref{fig:training_curve_375m}, both techniques lead to smoother training dynamics, with lower variance and fewer spikes, and achieve a lower final loss value.

\begin{table*}[!tb]
  \centering
  \caption{\small{\textbf{Image reconstruction results on COCO2017 and ImageNet-1k ($256\times 256$).}}
  }
  \label{tab:image_reconstruction}  
  \vspace{-10pt}
  \tablestyle{2pt}{1.05}
  \resizebox{1.02\textwidth}{!}{
  \begin{tabular}{lHcccrrrrrrrrrr}
  \toprule
  \multicolumn{7}{c}{} & \multicolumn{4}{c}{COCO2017 val} & \multicolumn{4}{c}{ImageNet-1k val} \\
    \cmidrule(r){8-11} \cmidrule(r){12-15} 
  Method & Data & Arch. & Quant. & Param. & \#bits & TP$_\uparrow$ & PSNR$_\uparrow$ & SSIM$_\uparrow$ & LPIPS$_\downarrow$ & rFID$_\downarrow$ & PSNR$_\uparrow$ & SSIM$_\uparrow$ & LPIPS$_\downarrow$ & rFID$_\downarrow$  \\
  \midrule
  \scriptsize{DALL-E dVAE}~\cite{ramesh2021dalle} & CC+YF & C & VQ & 98M & 13 & 34.0 & 25.15\std{3.49} & .7497\std{.1124} & .3014\std{.1221} & 55.07 & 25.46\std{3.93} & .7385\std{.1343} & .3127\std{.1480} & 36.84 \\  
  MaskGIT~\cite{chang2022maskgit}  & IN-1k & C & VQ & 54M & 10 & 37.6 & 17.52\std{2.75} & .4194\std{.1619} & .2057\std{.0473} & 8.90 & 17.93\std{2.93} & .4223\std{.1827} & .2018\std{.0543} & 2.23 \\
  SD-VAE 1.x~\cite{rombach2022ldm} & OImg & C & VQ & 68M & 14 & 22.4 & 22.54\std{3.55} & .6470\std{.1409} & .0905\std{.0323} & 6.07 & 22.82\std{3.97} & .6354\std{.1644} & .0912\std{.0390} & 1.23 \\
  ViT-VQGAN~\cite{yu2022vitvqgan} & IN-1k & T & VQ & 182M & 13 & $^\dagger$7.5 & - & - & - & - & - & - & - & 1.55 \\ 
  BSQ-ViT~\cite{zhao2025bsq} & IN-1k & T & BSQ & 174M & 18 & 45.1 & 25.08\std{3.57} & .7662\std{.0993} & .0744\std{.0295} & 5.81  & 25.36\std{4.02} & .7578\std{.1163} & .0761\std{.0358} & 1.14 \\
  \LSQ-ViT & IN-1k & T & \LSQ & 174M & $\lessapprox$18 & 45.1 & {\bf 26.00}\std{3.67} & {\bf .8008}\std{.0879} & {\bf .0632}\std{.0262} & {\bf 5.15} & {\bf26.37}\std{4.15} & {\bf.7934}\std{.1011} & {\bf.0622}\std{.0317} & {\bf0.83} \\
  \bottomrule
  \end{tabular}
  }
  \vspace{-10pt}
\end{table*}

\subsection{Sampling}
\label{sec:generation:sampling}

The sampling follows convention~\cite{tian2024var}.
We apply classifier-free guidance (CFG), first proposed for diffusion models ~\citep{ho2022cfg} and later adopted in AR-based models~\citep{mentzer2023fsq,sun2024llamagen}.
At inference, each token's logit $\vz_g$ is formed by $\vz_g = \vz_u + s(\vz_c - \vz_u)$, where $\vz_c$ is the conditional logit, $\vz_u$ is the unconditional logit, and $s$ is the CFG scale.
We use layer-wise linearly scaling CFG, first introduced in Infinity~\citep{han2025infinity}.
We observe that linearly scaling top-$K$ is also helpful.
For the factorized $d$-itwise configuration, we apply CFG on the normalized probability, \ie $p_g = p_u + s(p_c - p_u)$, where $p_{\cdot} = \exp
\left( \sum_{i}^d \log p_{\cdot}(\vc^{(i)}) \right)$ according to \cref{eq:factorized}.
Nucleus sampling (top $p$)~\cite{holtzman2019nucleus} is also used.

\section{Experiments}
\label{sec:exp}

\subsection{Experimental setup}

\myparagraph{Architectures.}
We train the image tokenizer with different quantization methods on ImageNet-1K~\citep{russakovsky2015imagenet}.
The experiments cover two network architectures: (1) Vision Transformers (ViT) \citep{dosovitskiy2021vit}, which runs at a high throughput and yields high reconstruction fidelity, 
and (2) ConvNets, which are more commonly seen in image generation.
We compare our method with VQ-based~\citep{rombach2022ldm,tian2024var} and BSQ-based methods~\citep{han2025infinity}.
The training details are specified in the Appendix.

\myparagraph{Training objectives.}
We use a weighted average of three losses, the mean absolute error (MAE, $\ell_1$), GAN, and perceptual loss, \emph{without} any other regularization terms.
The MAE, GAN, and perceptual loss optimize the PSNR, FID, and LPIPS score according to their respective definitions.
Therefore, this trio can no longer be simplified.

\myparagraph{Evaluation.}
We evaluate image compression in the Kodak Lossless True Color Image Suite.
It includes 24 24-bit lossless color PNG images.
We report PSNR and
MS-SSIM~\cite{wang2003msssim} at different levels of bits per pixel (BPP).
We assess image reconstruction and generation on the ImageNet-1k validation set.
Reconstruction is measured by FID, PSNR, SSIM, and LPIPS~\cite{zhang2018lpips}.
Generation is measured by FID, Inception Score (IS)~\cite{salimans2016is}, and improved precision
and recall (IPR)~\cite{kynkaanniemi2019ipr}, calculated by the \textit{ADM Tensorflow Evaluation Suite}~\cite{dhariwal2021adm}.
We use rFID/gFID to disambiguate.

\subsection{Main results: Comparison to state-of-the-art}
\label{sec:exp:main_result}

\vspace{-0.02in}
\myparagraph{State-of-the-art image reconstruction.}
\label{sec:exp:reconstruction}
\Cref{tab:image_reconstruction} compares the image reconstruction results on COCO 2017 and ImageNet-1k.
A ViT-based auto-encoder with \LSQ reduces rFID by $10\mathord\sim20\mathord\%$ and improves all other metrics.

\begin{SCtable}[1][!tb]
    \centering
    \caption{
    \small{\textbf{Image compression on Kodak.}
    \LSQs$^{\dagger}$ use only $\ell_1$ loss and does \textit{not} use arithmetic coding.
    }}
   \label{tab:compression_kodak}
    \tablestyle{2pt}{1.05}
    \begin{tabular}{lcccHH}
    \toprule
    Method & BPP$_\downarrow$ & PSNR$_\uparrow$ & MS-SSIM$_\uparrow$ & & LPIPS$_\downarrow$ \\
    \midrule
    JPEG2000 & 0.2986 & 29.192 & .9304 & (11.574) &  .1892 \\
    WebP & 0.2963 & 29.151 & .9396 & (12.193) & .1655 \\
    MAGVIT2 & 0.2812 & 23.467 & .8452 & (8.103) & .1260 \\
    BSQViT  & 0.2812 & 27.785 & .9481 & (12.852) & .0823 \\
    \LSQs$^{\dagger}$ & {\bf 0.2747} & {\bf 29.632} & {\bf .9637} & (14.398) & .1842 \\
    \bottomrule
    \end{tabular}
\vspace{-0.05in}
\end{SCtable}

\myparagraph{State-of-the-art image compression.}
\label{sec:exp:compression}
We show the compression results on Kodak in~\Cref{tab:compression_kodak}.
Since the resolution is $768\times512$ or $512\times768$, we encode/decode them in $256\times256$ tiles without overlapping or padding.
We compare our method with traditional codecs, including JPEG2000~\citep{openjpeg} and WebP~\citep{google2025webp}, and tokenizer-based approaches, including MAGVITv2~\citep{yu2023magvit2} and BSQViT~\citep{zhao2025bsq}.
\LSQ-ViT achieves higher PSNR and MS-SSIM scores while using a slightly smaller BPP.
Note that the rate-distortion tradeoff can be further improved ($\mathord\sim25\%$ less bitrate) by training an unconditional AR model for arithmetic coding~\cite{deletang2024lmic,zhao2025bsq}, which is not the primary focus of this paper.

\begin{SCfigure}[1][!b]
  \vspace{-30pt}
  \centering
  \includegraphics[width=0.5\linewidth]{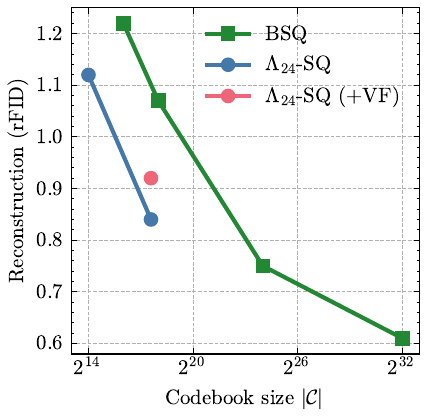}
  \caption{
  \small{\textbf{VAR Tokenizer.}}
  VAR+\LSQ achieves an rFID of 0.84; VAR+\LSQ (+VF) achieves an rFID of 0.92.
  More metrics are given in \Cref{tab:var_tokenizer}.
  }
  \label{fig:var_tokenizer}
\end{SCfigure}

\myparagraph{State-of-the-art image generation.}
\label{sec:exp:generation}
We select the class-conditioned Infinity~\cite{han2025infinity} as a baseline.
Infinity-CC employs a 7-level next-scale prediction backbone, saving $\mathord\sim25\%$ tokens and running $\mathord\sim30\%$ faster than VAR (10 levels)~\cite{tian2024var}.

\begin{table}[!tb]
    \centering
    \caption{
    \small{\textbf{Image generation on ImagetNet.}}
    $^{(re)}$ refers to rejection sampling.
    }
    \label{tab:image_generation}
    \vspace{-10pt}
    \tablestyle{2pt}{1.05}
    \begin{tabular}{l|cccc|lr}
    \toprule
    Method  & gFID$_\downarrow$ & IS$_\uparrow$  & Prec$_\uparrow$ & Rec$_\uparrow$ & \# Params & Steps \\
    \midrule
    VQGAN$^{(re)}$~\cite{esser2021vqgan} & 5.20 & 280.3 & - & - &  1.4B & 256 \\
    VIM$^{(re)}$~\cite{yu2022vitvqgan} & 3.04 & 227.4 & - & - & 1.7B & 1024 \\
    RQ-TF$^{(re)}$~\cite{lee2022rq} & 3.80 & 323.7 & - & - & 3.8B & 68 \\
    LlamaGen~\cite{sun2024llamagen} & 3.05 & 222.3 & 0.80 & 0.58 & 3.1B & 256 \\
    \midrule
    VAR-$d24$~\cite{tian2024var}  & 2.09 & 312.9 & 0.82 & 0.59 & 1.0B & 10 \\
    $\infty$-CC+\LSQ & 2.18 & {332.3} & 0.78 & {0.63} & 1.0B (0.3B) & 7 \\
    \midrule
    VAR-$d30$~\cite{tian2024var}  & 1.92 & 323.1 & 0.82 & 0.59 & 2.0B & 10 \\
    $\infty$-CC+\LSQ & 1.82 & 333.4 & 0.78 & 0.64 & 2.8B (0.4B) & 7 \\
    \rowcolor{gray!30}
    {(Val data)} & {1.78} & {236.9} & {0.75} & {0.67} & {-} & {-} \\
    \bottomrule
    \end{tabular}
\end{table}

\noindent\textit{(i) VAR tokenizer.}
We train a VAR tokenizer with the standard schedule (100 epochs) suggested in Infinity~\cite{han2025infinity}.
We use \LSQ as the bottleneck with two codebook sizes: (1) a complete codebook, whose bitrate is similar to BSQ ($d=18$), and (2) a subset of 16,384 codes, whose bitrate is equivalent to BSQ ($d=14$).
\Cref{fig:var_tokenizer} clearly demonstrates the superiority of our method.

\noindent\textit{(ii) VAR generation.}
\Cref{tab:image_generation} shows the generation results on ImageNet-1k.
We also provide the oracle result computed from the validation set in the bottom row.
Infinity-CC+\LSQ works comparably with VAR-$d24$ in terms of model size (1B) while being 30\% more efficient.
Note that our results achieve a higher recall and push the precision-recall tradeoff closer to the validation oracle.
We attribute this to the larger codebook, which better captures visual diversity.
When the parameters increase to 2.8B, $\infty$-CC+\LSQ achieves an FID of 1.82, which is comparable to both VAR-$d30$ and the oracle result.

\begin{table}[!tb]
    \centering
    \caption{\small{
    \textbf{Quantizer with higher $\delta_\mathrm{min}$ lead to better reconstruction.}
    VQ ($\mathcal{PN}$-): projected normal distribution initialization.}
    Note that we gray out the last two rows to indicate that the learnable configurations are \textbf{not} used elsewhere.
    }
    \label{tab:npq_ablation}
    \vspace{-10pt}
    \resizebox{1.0\linewidth}{!}{
    \tablestyle{2pt}{1.05}
    \begin{tabular}{ccc|ccccc}
    \toprule
        & Method & $|\mathcal{C}|$ & Utility$_\uparrow$ & rFID$_\downarrow$ & LPIPS$_\downarrow$ & SSIM$_\uparrow$ & PSNR$_\uparrow$ \\
    \midrule
    \multirow{8}{*}{(Fixed)} & VQ (RP, $U$) & ${2^{14}}$ & 95.09\% & 13.08 & 0.1080 & 0.7086 & 23.018 \\
    & VQ (RP, $\mathcal{N}$) & ${2^{14}}$ & 100.0\% & 12.00 & 0.1021 & 0.7240 & 23.354 \\
    & BSQ & ${2^{14}}$ & 97.21\% & 12.98 & 0.1048 & 0.7058 & 23.171 \\
    & \LSQ & ${2^{14}}$ & 100.0\% & {\bf 11.16} & {\bf 0.1007} & {\bf 0.7258} & {\bf 23.390} \\
    \cmidrule{2-8}
    & BSQ & ${2^{17}}$ & 57.04\% & {12.46} & 0.0963 & 0.7296 & 23.742 \\
    & BSQ & ${2^{18}}$ & 70.00\% & {10.96} & 0.0914 & 0.7351 &  23.752 \\
    & VQ (RP, $\mathcal{N}$) & $196,560$ & 94.87\% & ~~9.10 & 0.0829 & 0.7624 & 24.216 \\
    & \LSQ & ${196,560}$ & 94.84\% & {\bf ~~8.98} & {\bf 0.0811} & {\bf 0.7647} & {\bf 24.282} \\
    \hline
    \hline
    \rowcolor{gray!30}
     & VQ ($\mathcal{PN}$-init.) & $196,560$ & 95.45\% & ~~9.13 & 0.0832 & 0.7623 & 24.226 \\
    \rowcolor{gray!30}
    \multirow{-2}{*}{(Learn)} & \LSQ & $196,560$ & 94.78\% & ~~8.78 & 0.0820 & 0.7644 & 24.274 \\
    \bottomrule
    \end{tabular}
    }
\end{table}

\subsection{Scientific investigations and ablative studies}

\myparagraph{Dispersiveness leads to better rate-distortion tradeoff.}
We start from the comparison in \Cref{fig:delta_vs_N} and ask if \textit{dispersiveness, quantified by higher $\delta_\mathrm{min}(N)$, leads to better rate-distortion tradeoff for visual tokenization}.
We train a plain ViT-small encoder-decoder on ImageNet-$128\times128$ while varying only the quantization bottleneck.
We also test two vocabulary sizes, medium ($2^{14}$) and large ($196,560$).
With the codebook fixed, we find \LSQ achieves the best rFID, LPIPS, SSIM, and PSNR.
The random projection VQ and BSQ follow behind.
We also test learnable codebooks given these as initialization.
The conclusion still holds, and a learnable codebook does \textit{not} greatly affect the final results.

\begin{figure}[!tb]
    \centering
    \includegraphics[width=0.9\linewidth]{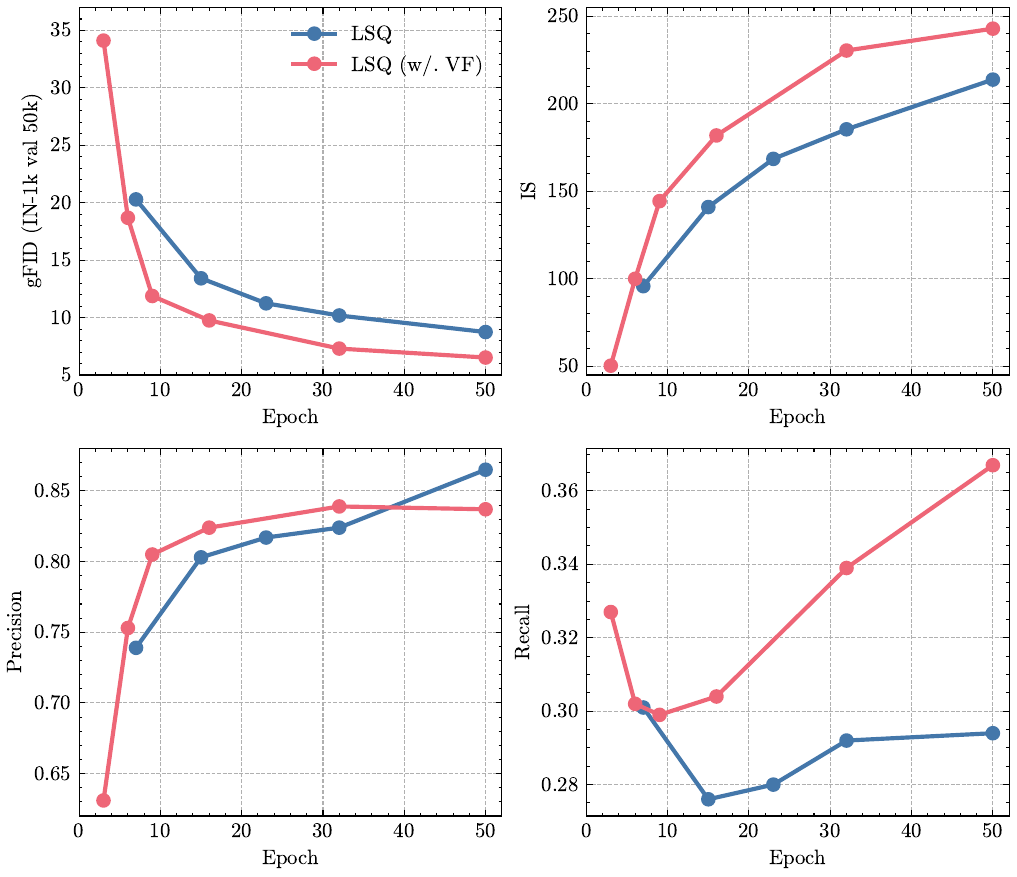}
    \vspace{-10pt}
    \caption{\small{\textbf{VF alignment improves convergence and final generation results, especially recall.}}
    The model has 12 layers (240M).
    }
    \label{fig:var_generation_curve}
\end{figure}

\myparagraph{VF alignment helps \textit{discrete} tokens, too.}
\Cref{fig:var_tokenizer} shows a worse reconstruction quality after aligning with the DINOv2 feature~\cite{oquab2023dinov2}.
However, \Cref{fig:var_generation_curve} demonstrates that the VAR generation using a VF-aligned tokenizer converges faster and achieves better final results in gFID, IS, and especially recall.
This extends the findings in VAVAE~\cite{yao2025vavae} about VF alignment
from \textit{continuous} latents to \textit{discrete} ones. 

\myparagraph{VAR generation with different heads.}
We test various prediction heads, namely $18\times$ BCE \vs $262,144$-way CE for $\infty$-CC+BSQ, $24\times$ 9-way CE \vs $196,560$-way CE for $\infty$-CC+\LSQ in \Cref{tab:var_generator_heads}, showing that \LSQ+CE achieves great results despite simplicity.
The factorized $d$-itwise prediction yields worse gFID and lower recall in both cases, implying that \textit{factorized approximation sacrifices diversity}. 
We also find that the optimal sampling hyperparameters vary and conduct a small-scale grid search in \Cref{fig:grid_search}.

\begin{SCtable}[1][!tb]
    \centering
    \resizebox{0.66\linewidth}{!}{
    \tablestyle{2pt}{1.05}
    \begin{tabular}{cc|Hcc|cccc}
    \toprule
        & Pred. head & top $k$ & gFID$_\downarrow$ & IS$_\uparrow$ & Prec$_\uparrow$ & Rec$_\uparrow$ \\
    \midrule
        BSQ~\citep{han2025infinity} & BCE & 900 & 10.7 & 219.6 & 0.85 & 0.21 \\
        BSQ~\citep{han2025infinity} & CE & 10,000 & 10.3 & 187.3 &  0.85 & 0.27 \\
        \LSQ & 9-way CE & 250 & 11.7 & 155.8 & 0.82 & 0.29 \\
        \LSQ & CE & 10,000 & 8.7 & 215.4 & 0.85 & 0.30 \\ 
    \bottomrule
    \end{tabular}
    }
    \hspace{-8pt}
    \caption{
    \small{\textbf{$\infty$-CC with different prediction heads.}
    $\mathrm{CFG}=2$, $p=0.95$, and $K$ varies.
    Grid search in \Cref{fig:grid_search}.
    }}
    \label{tab:var_generator_heads}
\end{SCtable}

\begin{figure}[!tb]
    \centering
    \includegraphics[width=0.9\linewidth]{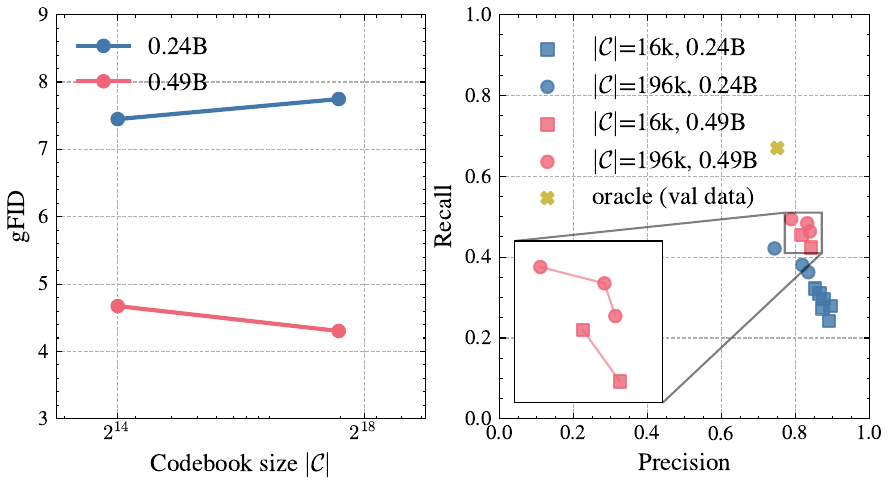}
    \vspace{-10pt}
    \caption{\small{\textbf{Scaling effect of the codebook size.}}
    \textbf{Left}: Increasing the codebook size improves gFID when the model is large (0.49B).
    \textbf{Right}: Increasing the codebook size pushes the Precision-Recall Pareto frontier towards the oracle precision-recall derived from the validation set (see the zoom-in at the bottom left).
    }
    \label{fig:scaling_codebook_size}
\end{figure}

\myparagraph{Scaling the codebook size \emph{does} matter.}
The last but not least critical question is \emph{whether increasing the codebook size benefits the generation results}.
To answer this, we used the two VAR tokenizers with $\mathcal{C}=196,560$ \vs $16,384$) with reconstruction results in \Cref{fig:var_tokenizer}, and trained VAR models with varied sizes on top while keeping all the rest settings the same.
To report the gFID, we search the sampling hyperparameters to find an optimal value.
From \Cref{fig:scaling_codebook_size}, we conclude that increasing the codebook size improves gFID when the model is large,~\eg, 12-layer (0.24B) to 16-layer (0.49B).
This echoes the finding in LLMs that larger models deserve larger vocabularies~\cite{tao2024scaling}.
We look at the improved precision and recall metric in the right subplot of \Cref{fig:scaling_codebook_size}.
We use top-$p=0.95$, CFG of $\mathrm{lin}(1, 0.33)$, and vary top-$k$ to obtain the data points.
We find that when the codebook size increases, the precision-recall Pareto frontier moves towards the oracle precision-recall derived from the val set.

\section{Related work}

\myparagraph{Vector quantization (VQ)}~\citep{gray1984vq,van2017vqvae} lays the foundations of learning discrete visual tokens.
However, VQ is notoriously difficult to train, which is attributed to the misalignment between the embedding distribution of the model and the codebook~\cite{huh2023straightening}.
Optimization tricks are then introduced,~\eg Gumbel Softmax~\cite{jang2017gumbelsoftmax,baevski2020gumbelvq}, Rotation trick~\citep{fifty2025rotation}, and Index Backpropagation~\cite{shi2025ibq}.
The line of lattice-based quantization methods covered in this paper~\citep{yu2023magvit2,mentzer2023fsq,zhao2025bsq} addresses this misalignment issue by keeping the codebook fixed.
Our \LSQ is an intuitive extension in this direction.

\myparagraph{Scaling visual tokenizers} has recently attracted attention and covers many directions, including
increasing parameters~\cite{xiong2025gigatok}, training data~\cite{hansen2025learnings}, unifying generation and understanding~\cite{zhao2025qlip,ma2025unitok}, and encoding multiple modalities~\cite{wang2024omnitokenizer,lu2025atoken}.
Our paper focuses on scaling the vocabulary size.
Despite several advances in reconstruction~\cite{zhu2024vqganlc,shi2025ibq}, none have reported that an expanded vocabulary benefits generation yet.
Our paper shows that a visual autoregressive model scales to a very large codebook ($\mathord\sim200K$) without tricks such as index subgrouping~\cite{yu2023magvit2},~\etc.

\myparagraph{Autoregressive visual generation} applies an autoregressive model to visual generation~\cite{efros1999texture,van2016pixelcnn,chen2020igpt} similar to the LLM paradigm~\cite{bengio2003lm,radford2018gpt,brown2020gpt3}.
Modern AR model employs a visual tokenizer for efficiency~\cite{esser2021vqgan}.
Subsequent work explores what to auto-regress~\cite{yu2024titok,tian2024var} and auto-regressive order~\cite{yu2025rar,pang2025randar}.
Most AR models are based on a medium-sized visual vocabulary ($1K\mathord\sim\mathord10K$), limiting their potential~\cite{yu2023magvit2}.

\myparagraph{Lattice coding} has wide applications in digital communication \cite{zamir2014lattice} and cryptography \cite{peikert2016decade}.
In this paper, we borrow this concept to describe different non-parametric quantization methods and others in the same language.
This further inspires us to devise new quantization methods based on the principle of \textit{densest hypersphere packing}.
The hyperspherical prior is also loosely related to some recent work about learning on a spherical manifold~\cite{davidson2018svae,loshchilov2025ngpt}.

\section{Conclusion}

We have introduced spherical Leech quantization (\LSQ), a novel quantization method that scales the visual codebook to $\mathord\sim200K$, and demonstrated its applications in visual compression, reconstruction, and generation on ImageNet.
In the future, we are interested in verifying its effectiveness in larger-scale settings,~\eg, text-conditioned visual generation.

\myparagraph{Acknowledgments.}
This material is based upon work in part supported by the National Science Foundation under Grant
No. IIS-1845485.
The authors acknowledge the IFML Center for Generative AI and the Texas Advanced Computing Center (TACC) at The University of Texas at Austin for providing computational resources that have contributed to the research results reported within this paper.
This work was supported by a Hoffman-Yee Research Grant from the Stanford Institute for Human-Centered Artificial Intelligence (HAI).
This research was also supported in part by Lambda, Inc.
YZ would like to thank Jeffrey Ouyang-Zhang and Mi Luo for their help in setting up environments on TACC; Yi Jiang and Bin Yan for their clarification on VAR and Infinity baselines.

{
    \small
    \bibliographystyle{ieeenat_fullname}
    \bibliography{main}
}

\appendix
\clearpage
\setcounter{page}{1}

\section{Constructing Fibonacci Lattices}
\label{sec:appendix:fibonacci}

\myparagraph{Fibonacci lattice} constructs points that \textit{``are evenly distributed with each of them representing almost the same area"}~\citep{gonzalez2010measurement} in a unit square $[0,1)^2$
using the formula:
\begin{align}
    (x_i, y_i) = (i - \lfloor \frac{i}{\psi} \rfloor, \frac{i}{n}) \text{ for } 0\leq i < n, 
\end{align}
where $\psi =\lim_{n\rightarrow\infty}\left( \frac{F_{n+1}}{F_{n}} \right) = \frac{1+\sqrt{5}}{2}$.
We can map this point distribution to a unit-length sphere $\sS^{2}$ using cylindrical equal-area projection.
\begin{align}
    \begin{pmatrix} x_i \\ y_i \end{pmatrix} \rightarrow
    \begin{pmatrix*}[l] \theta_i = 2\pi x_i \\ \phi_i = \arccos(1-2y_i) \end{pmatrix*} \rightarrow
    \begin{pmatrix*}[l] x'_i = \cos\theta_i\sin\phi_i \\ y'_i = \sin\theta_i\sin\phi_i \\ z'_i = \cos\phi_i \end{pmatrix*}.
\end{align}

\subsection{Generalizing the Spherical Fibonacci Lattice to higher dimensions}

We provide the details to generalize the 3D  spherical Fibonacci lattice~\citep{gonzalez2010measurement} to higher dimensions~\citep{stackexchange3297830}.

Given a $d$-dimensional vector $\vu = (u_1, u_2, \cdots, u_{d+1}) \in \sS^{d} \subset \R^{d+1}$, we can also represent it in the hyperspherical coordinate system $(r, \varphi_1, \varphi_2, \cdots, \varphi_{d})$, where $\varphi_1\in[0, 2\pi]$ , $\varphi_2,\cdots,\varphi_{d}\in[0, \pi]$, and specifically, $r=1$ for $|\vu|=1$.
The conversion to Cartesian coordinates is given as follows:
\begin{align*}
u_{d+1} &= \cos(\varphi_{d}) \\
u_d &= \sin(\varphi_{d})\cos(\varphi_{d-1}) \\
\vdots \\
u_{2}   &= \sin(\varphi_{d})\sin(\varphi_{d-1})\cdots\sin(\varphi_{2})\cos(\varphi_{1}) \\
u_{1} &= \sin(\varphi_{d})\sin(\varphi_{d-1})\cdots\sin(\varphi_{2})\sin(\varphi_{1}).
\end{align*}

We examine the distribution over the angular coordinates $\rvphi_{1,\cdots,d-1,d}\in [0, 2\pi] \times [0,\pi]^{d-1}$.
\begin{align}
    p(\rvphi_{1:d}) &= p(\rvphi_1,\rvphi_2,\cdots,\rvphi_d) \\
    &= \rho(\rvphi_1)\rho(\rvphi_2|\rvphi_1)\cdots\rho(\rvphi_d|\rvphi_1,\cdots,\rvphi_{1:d-1}).
    \label{eq:angular_prob}
\end{align}

The key observation is that the angles are independently distributed.
To see this, if we fix $(\varphi_1, \cdots, \varphi_k)$, then $(\varphi_{k+1}, \cdots, \varphi_{d})$ parameterizes a ``subsphere" isoporphic to $\sS^{d-k}$ with a rescaled radius $r'=\sin(\varphi_1)\cdots\sin(\varphi_k)$.
In other words, $ p(\rvphi_{k+1:d} | \rvphi_{1:k}) = p(\rvphi_{k+1:d}) $ for any $k\in[d]$.
As such, \Eqref{eq:angular_prob} can be simplified to:
\begin{align}
    p(\rvphi_{1:d}) = \prod_{\alpha=1}^d p_\alpha(\rvphi_\alpha).
    \label{eq:angular_prob_2}
\end{align}

The absolute value of the Jacobian determinant for the change of variables $(u_1, \cdots, u_{d+1}) \mapsto (r, \varphi_1, \cdots, \varphi_d)$ is
\begin{align}
    \left| \frac{\partial (u_1, \cdots, u_{d+1})}{\partial (r, \varphi_1, \cdots, \varphi_d)} \right| 
= r^d\prod_{k=2}^{d} \sin^{k-1}\varphi_k.
\end{align}

Therefore,~\Eqref{eq:angular_prob} reduces to
\begin{align}
    p(\rvphi_{1:d}) \propto \prod_{k=2}^{d} \sin^{k-1}\varphi_k.
\end{align}
With \Eqref{eq:angular_prob_2}, we have $p(\rvphi_k)\propto \sin^{k-1}\varphi_k$.
Then we get the normalization constants $Z_k = \int_0^\pi \sin^{k-1}\varphi d\varphi=\frac{\sqrt{\pi}\cdot \Gamma(k/2)}{\Gamma((k+1)/2)}$ for $k=2,\cdots,d$ and $Z_1=\int_0^{2\pi} d\varphi_1=2\pi$.
Finally, we arrive at
\begin{align}
    P(\rvphi_k=\varphi_k)=
    \begin{cases}
    \frac{1}{2\pi}, & k=1 \\
    \frac{1}{\sqrt{\pi}}\frac{\Gamma(\frac{k+1}{2})}{\Gamma(\frac{k}{2})} \sin^{k-1}\varphi_k, & k=2,\cdots, d
    \end{cases}
\end{align}

Denote the cumulative distribution function with another variable $Y$
\begin{align}
    P(Y=y) &= F_{\rvphi}(\varphi) = \int_{0}^{\varphi_k} p(\rvphi=u)du \\
    &= 
    \begin{cases}
        \frac{1}{2\pi}y, & k=1 \\
        ...  \\
    \end{cases}.
\end{align}

The Fibonacci-like spiral ${\mY}^{(n)}=\left( \mY_1^{(n)}, \cdots, \mY_d^{(n)} \right)$ is generated by the following formula:
\begin{align}
    \mY_d^{(n)} &= \frac{n}{N+1}, \\
    \mY_{d-1}^{(n)} &= \{na_1\}, \\
    \vdots \\
    \mY_1^{(n)}  &= \{na_{d-1}\}, \\
\end{align}
where $\{x\} $ refers to $x$'s decimal part,~\ie $\{ x \} = x - \lfloor  x \rfloor $.
$a_{1:d}$ satisfies $\frac{a_i}{a_j} \notin \sQ, \forall i\neq j$.

The angles are given by taking the inverse:
\begin{align}
    \varphi_d ^{(n)}= F^{-1}(Y_d^{(n)})
\end{align}

\begin{align}
\varphi[t+1] = \varphi[t] - \frac{F(\varphi[t])-Y}{F'(\varphi[t])}
\end{align}

\begin{table*}[!tb]
    \caption{\small{\textbf{Generator matrix for the Leech lattice $\mathbbold{\Lambda}_{24}$}. The table is adapted from Table 4.12 in \cite{conway2013sphere}.}}
    \label{tab:generator_matrix_lattice}
    \begin{align*}
    \frac{1}{\sqrt{8}}
    \begin{bmatrix}
    \begin{array}{cccc|cccc|cccc|cccc|cccc|cccc}
    8 & 0 & 0 & 0 & 0 & 0 & 0 & 0 & 0 & 0 & 0 & 0 & 0 & 0 & 0 & 0 & 0 & 0 & 0 & 0 & 0 & 0 & 0 & 0 \\
    4 & 4 & 0 & 0 & 0 & 0 & 0 & 0 & 0 & 0 & 0 & 0 & 0 & 0 & 0 & 0 & 0 & 0 & 0 & 0 & 0 & 0 & 0 & 0 \\
    4 & 0 & 4 & 0 & 0 & 0 & 0 & 0 & 0 & 0 & 0 & 0 & 0 & 0 & 0 & 0 & 0 & 0 & 0 & 0 & 0 & 0 & 0 & 0 \\
    4 & 0 & 0 & 4 & 0 & 0 & 0 & 0 & 0 & 0 & 0 & 0 & 0 & 0 & 0 & 0 & 0 & 0 & 0 & 0 & 0 & 0 & 0 & 0 \\
    \midrule
    4 & 0 & 0 & 0 & 4 & 0 & 0 & 0 & 0 & 0 & 0 & 0 & 0 & 0 & 0 & 0 & 0 & 0 & 0 & 0 & 0 & 0 & 0 & 0 \\
    4 & 0 & 0 & 0 & 0 & 4 & 0 & 0 & 0 & 0 & 0 & 0 & 0 & 0 & 0 & 0 & 0 & 0 & 0 & 0 & 0 & 0 & 0 & 0 \\
    4 & 0 & 0 & 0 & 0 & 0 & 4 & 0 & 0 & 0 & 0 & 0 & 0 & 0 & 0 & 0 & 0 & 0 & 0 & 0 & 0 & 0 & 0 & 0 \\
    2 & 2 & 2 & 2 & 2 & 2 & 2 & 2 & 0 & 0 & 0 & 0 & 0 & 0 & 0 & 0 & 0 & 0 & 0 & 0 & 0 & 0 & 0 & 0 \\
    \midrule
    4 & 0 & 0 & 0 & 0 & 0 & 0 & 0 & 4 & 0 & 0 & 0 & 0 & 0 & 0 & 0 & 0 & 0 & 0 & 0 & 0 & 0 & 0 & 0 \\
    4 & 0 & 0 & 0 & 0 & 0 & 0 & 0 & 0 & 4 & 0 & 0 & 0 & 0 & 0 & 0 & 0 & 0 & 0 & 0 & 0 & 0 & 0 & 0 \\
    4 & 0 & 0 & 0 & 0 & 0 & 0 & 0 & 0 & 0 & 4 & 0 & 0 & 0 & 0 & 0 & 0 & 0 & 0 & 0 & 0 & 0 & 0 & 0 \\
    2 & 2 & 2 & 2 & 0 & 0 & 0 & 0 & 2 & 2 & 2 & 2 & 0 & 0 & 0 & 0 & 0 & 0 & 0 & 0 & 0 & 0 & 0 & 0 \\
    \midrule
    4 & 0 & 0 & 0 & 0 & 0 & 0 & 0 & 0 & 0 & 0 & 0 & 4 & 0 & 0 & 0 & 0 & 0 & 0 & 0 & 0 & 0 & 0 & 0 \\
    2 & 2 & 0 & 0 & 2 & 2 & 0 & 0 & 2 & 2 & 0 & 0 & 2 & 2 & 0 & 0 & 0 & 0 & 0 & 0 & 0 & 0 & 0 & 0 \\
    2 & 0 & 2 & 0 & 2 & 0 & 2 & 0 & 2 & 0 & 2 & 0 & 2 & 0 & 2 & 0 & 0 & 0 & 0 & 0 & 0 & 0 & 0 & 0 \\
    2 & 0 & 0 & 2 & 2 & 0 & 0 & 2 & 2 & 0 & 0 & 2 & 2 & 0 & 0 & 2 & 0 & 0 & 0 & 0 & 0 & 0 & 0 & 0 \\
    \midrule
    4 & 0 & 0 & 0 & 0 & 0 & 0 & 0 & 0 & 0 & 0 & 0 & 0 & 0 & 0 & 0 & 4 & 0 & 0 & 0 & 0 & 0 & 0 & 0 \\
    2 & 0 & 2 & 0 & 2 & 0 & 0 & 2 & 2 & 2 & 0 & 0 & 0 & 0 & 0 & 0 & 2 & 2 & 0 & 0 & 0 & 0 & 0 & 0 \\
    2 & 0 & 0 & 2 & 2 & 2 & 0 & 0 & 2 & 0 & 2 & 0 & 0 & 0 & 0 & 0 & 2 & 0 & 2 & 0 & 0 & 0 & 0 & 0 \\
    2 & 2 & 0 & 0 & 2 & 0 & 2 & 0 & 2 & 0 & 0 & 2 & 0 & 0 & 0 & 0 & 2 & 0 & 0 & 2 & 0 & 0 & 0 & 0 \\
    \midrule
    0 & 2 & 2 & 2 & 2 & 0 & 0 & 0 & 2 & 0 & 0 & 0 & 2 & 0 & 0 & 0 & 2 & 0 & 0 & 0 & 2 & 0 & 0 & 0 \\
    0 & 0 & 0 & 0 & 0 & 0 & 0 & 0 & 2 & 2 & 0 & 0 & 2 & 2 & 0 & 0 & 2 & 2 & 0 & 0 & 2 & 2 & 0 & 0 \\
    0 & 0 & 0 & 0 & 0 & 0 & 0 & 0 & 2 & 0 & 2 & 0 & 2 & 0 & 2 & 0 & 2 & 0 & 2 & 0 & 2 & 0 & 2 & 0 \\
   -3 & 1 & 1 & 1 & 1 & 1 & 1 & 1 & 1 & 1 & 1 & 1 & 1 & 1 & 1 & 1 & 1 & 1 & 1 & 1 & 1 & 1 & 1 & 1 \\
    \end{array}
    \end{bmatrix}
    \end{align*}
\end{table*}

\section{Details of the Leech Lattice}
\label{sec:appendix:leech}

\myparagraph{Generator matrix.}
The generator matrix for the \textit{unconstrained} $\mathbbold{\Lambda}_{24}$ is given in \Cref{tab:generator_matrix_lattice}.

\section{Experimental Details}

\subsection{Architectures}

\myparagraph{ViT tokenizer.}
The architecture used to get the main results follows \cite{zhao2025bsq}.
To conduct ablative studies in \Cref{tab:npq_ablation}, we use a smaller architecture (ViT-small) for fast iteration.
Both the hidden dimension and the number of layers are halved.

\myparagraph{CNN-based VAR tokenizer.}
The architecture follows Infinity~\cite{han2025infinity}\footnote{\url{https://github.com/FoundationVision/BitVAE}}.
We use 7 scales: $[1^2, 2^2, 4^2, 6^2, 8^2, 12^2, 16^2]$, which amounts to 521 tokens in total.

\myparagraph{Infinity-CC.}
The Infinity model used in the paper mostly follows the original Infinity paper~\cite{han2025infinity}.
We make two key changes: (1) The text condition is converted to class condition, which is representation by a single \texttt{<SOS>} token.
The vocabulary size is augmented to $|\mathcal{C}| + K + 1$, where $K=1000$.
The additional $1$ refers to the ``no-class'' index, which is used to randomly replace the original class index with a probability of $0.1$, to enable classifier-free guidance (CFG) at the sampling phase.
(2) We revert the shared AdaLN to an unshared version following VAR~\cite{tian2024var}.
\Cref{tab:infinity_cc_config} summarizes the model configurations and parameter size in this paper. 

\subsection{Training specifications}
\myparagraph{ViT tokenizer.}
We train the image tokenizer with a batch size of 32 per GPU.
We use AdamW optimizer with $(\beta_1, \beta_2) = (0.9, 0.99)$ with $1\times10^{-4}$ weight decay.
The base learning rate is $4\times10^{-7}$ (or a total learning rate of $1\times10^{-4}$) and follows a half-period cosine
annealing schedule.
The model is trained for 1M steps, which amounts to 200 epochs over the entire
ImageNet-1k training set.
We use an $\ell_1$ loss weight of $1$, a perceptual loss weight of $0.1$, and an adversarial loss weight of $0.1$ throughout the experiments.

\myparagraph{VAR tokenizer.}
We train the VAR tokenizer with a batch size of 8 per GPU.
Two schedules are used: (1) the fast schedule trains the model for 500k iterations with 8 GPUs, which approximately sees the training data 25 epochs; (2) the standard schedule trains the model for 500k iterations with 32 GPUs, which is approximately 100 epochs.

\begin{table}[!tb]
    \centering
    \caption{\small{\textbf{Model configurations for Infinity-CC.}}}
    \label{tab:infinity_cc_config}
    \tablestyle{3pt}{1.05}
    \begin{tabular}{cccr|r}
    \toprule
        layer & embed dim & \# heads & \# params (head) &  \# epochs \\
    \midrule
        12  &  768  & 8  & 242M (151M) & 50 \\
        16  & 1152  & 12 & 394M (226M) & 200 \\
        24  & 1536  & 16 & 1B (300M)   & 350 \\
        32  & 2048  & 16 & 2.8B (402M) & 400 \\
    \bottomrule
    \end{tabular}
\end{table}

\section{More Results}

\subsection{VAR tokenization}
First, we retrain a VAR tokenizer with a fast schedule (25 epochs).
We use \LSQ as the bottleneck with two codebook sizes: (1) the full codebook, whose bitrate is similar to BSQ ($d=18$), and (2) a subset of 16,384 codes, whose bitrate is equivalent to BSQ ($d=14$).
From the upper half of \Cref{tab:var_tokenizer}, \LSQ outperforms BSQ in all metrics in both cases. 
Next, we train a VAR tokenizer with the standard schedule (100 epochs) suggested in Infinity~\cite{han2025infinity}.
The full numbers are reported in the bottom half of \Cref{tab:var_tokenizer}, supplementing \Cref{fig:var_tokenizer}.

\begin{table}[!tb]
    \centering
    \caption{\small{\textbf{VAR Tokenizer.}}}.
    \label{tab:var_tokenizer}
    \tablestyle{3pt}{1.05}
    \begin{tabular}{l|Hr|cccc}
    \toprule
                     &  $d$ & $| \mathcal{C} |$  & rFID$_\downarrow$ & LPIPS$_\downarrow$  & SSIM$_\uparrow$   & PSNR$_\uparrow$  \\
    \midrule
        \multicolumn{6}{l}{\textit{(fast schedule)}} \\
        BSQ &  14 & 16,384    & 1.82 & 0.1268 & 0.5626 & 19.989 \\
        \LSQs$^\dagger$ &  24  & 16,384 & 1.36 & 0.1170 & 0.5957  & 20.639 \\ 
        BSQ &  18  & 262,144 & 1.29 & 0.1106 & 0.6006 & 20.683 \\
        \LSQ &  24  & 196,560 & 1.08 & 0.1005 & 0.6280 & 21.315 \\ 
        \LSQ (vf) & 24 & 196,560 & 1.18 & 0.1088 & 0.6006 & 20.734 \\
        \hline
        \multicolumn{6}{l}{\textit{(standard schedule)}} \\
        BSQ &  18  & 262,144 & 1.07 & 0.1064 & 0.6035 & 20.430 \\
        \LSQ & 24 & 196,560 & {\bf 0.84} & {\bf 0.0954} & {\bf 0.6333} & {\bf 21.535} \\
        \LSQ (vf) & 24 & 196,560 & \underline{0.92} & \underline{0.1041} & \underline{0.6118} & \underline{21.188} \\
    \bottomrule
    \end{tabular}
\end{table}

\begin{figure*}[!tb]
    \centering
    \begin{subfigure}[t]{0.45\linewidth}
      \centering
        \includegraphics[width=\linewidth]{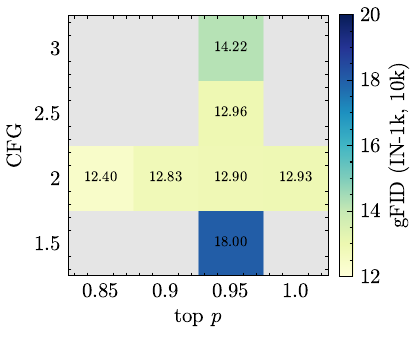}
        \caption{BSQ + BCE}
        \label{fig:grid_search_bsq_bce}
    \end{subfigure}%
    \begin{subfigure}[t]{0.45\linewidth}
      \centering
        \includegraphics[width=\linewidth]{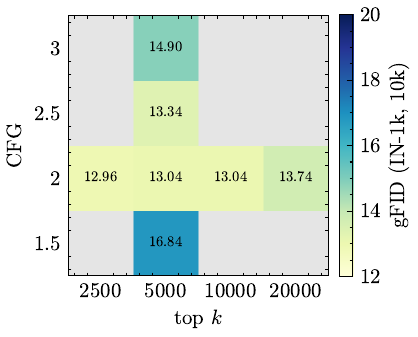}
        \caption{BSQ + CE}
        \label{fig:grid_search_bsq_cce}
    \end{subfigure}%
    \hfill
    \begin{subfigure}[t]{0.45\linewidth}
      \centering
        \includegraphics[width=\linewidth]{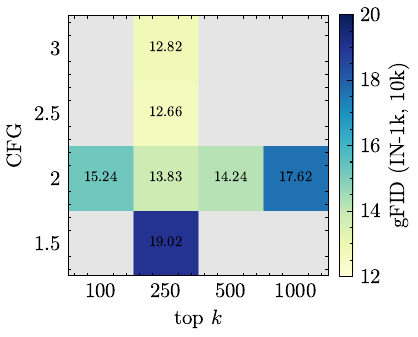}
        \caption{\LSQ + $d$-way CE}
        \label{fig:grid_search_lsq_bce}
    \end{subfigure}
    \begin{subfigure}[t]{0.45\linewidth}
      \centering
        \includegraphics[width=\linewidth]{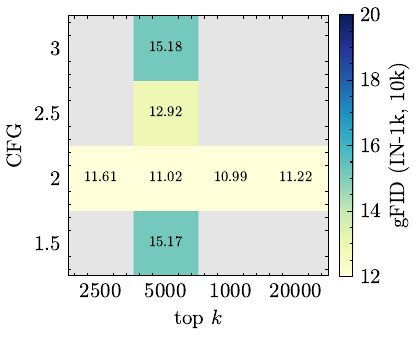}
        \caption{\LSQ + CE}
        \label{fig:grid_search_lsq_cce}
    \end{subfigure}
    \caption{\small{\textbf{Hyperparameter grid search supplementing \Cref{tab:var_generator_heads}.}} We use a fixed temperature $\tau=1$ in (a-d) and top $p=0.95$ in (b-d). }
    \label{fig:grid_search}
\end{figure*}

\subsection{VAR generation}

\myparagraph{Grid search of sampling parameters.}
We run a small-scale grid search of sampling hyperparameters for Infinity-CC with different prediction heads.
We compare the gFID score on IN-1k by generating 10 samples per class (10k generated samples in total).
From \Cref{fig:grid_search}, we conclude that the optimal top $k$ varies significantly across different prediction head settings.

\myparagraph{Advanced sampling techniques.}
In \Cref{sec:generation:sampling}, we introduced advanced sampling techniques, including layerwise linearly scaling CFG and linearly scaling top-$k$.
We show related ablation studies in \Cref{tab:advanced_sampling}.
We use $\mathrm{lin}(x_0,s)$ to denote the linear scaling strategy, which starts from $x_0$ and changes by $s$ per scale.
We can see that both layerwise linear scaling CFG and top-$k$ bring a noticeable improvement. 

It is also worth noting that, according to the bottom half of \Cref{tab:advanced_sampling}, the optimal $k$ decreases when the tokenizer is trained with the VF loss.
This is most likely because the probability density is more skewed, as is illustrated in \Cref{fig:codebook_histogram}. 

\begin{table}[!tb]
    \centering
    \caption{
    \small{\textbf{Advanced sampling techniques.}}
    $\mathrm{lin}(x_0,\pm s)$ denotes the linear scaling strategy which starts from $x_0$ and increment/decrements by $s$ per scale. 
    }.
    \label{tab:advanced_sampling}
    \tablestyle{3pt}{1.05}
    \begin{tabular}{ll||llc}
    \toprule
        Tokenizer & rFID & CFG & top $k$ & gFID \\
    \midrule
        \LSQ (25 ep) & 1.08 & 2  & $5\times10^3$  & 8.78 \\
        \LSQ (100 ep) & 0.84 & 2 & $5\times10^3$  & 7.46 \\
        \LSQ (100 ep) & 0.84 & $\mathrm{lin}(1,0.33)$ & $5\times10^3$ & 6.81 \\
        \LSQ (100 ep) & 0.84 & $\mathrm{lin}(1,0.25)$ & $5\times10^3$ & 7.33 \\
        \LSQ (100 ep) & 0.84 & $\mathrm{lin}(1,0.33)$ & $\mathrm{lin}(10^4,-10^3)$ &  6.68 \\
    \midrule
        \LSQ (vf)     & 1.18 & $\mathrm{lin}(1,0.33)$ & $5\times10^3$ & 5.79 \\
        \LSQ (vf)     & 1.18 & $\mathrm{lin}(1,0.33)$ & $2,500$ & 5.41 \\
        \LSQ (vf)     & 1.18 & $\mathrm{lin}(1,0.33)$ & $\mathrm{lin}(2000,-100)$  & 5.30 \\        
    \bottomrule
    \end{tabular}

\end{table}

\begin{figure*}[!tb]
    \centering
    \includegraphics[width=\linewidth]{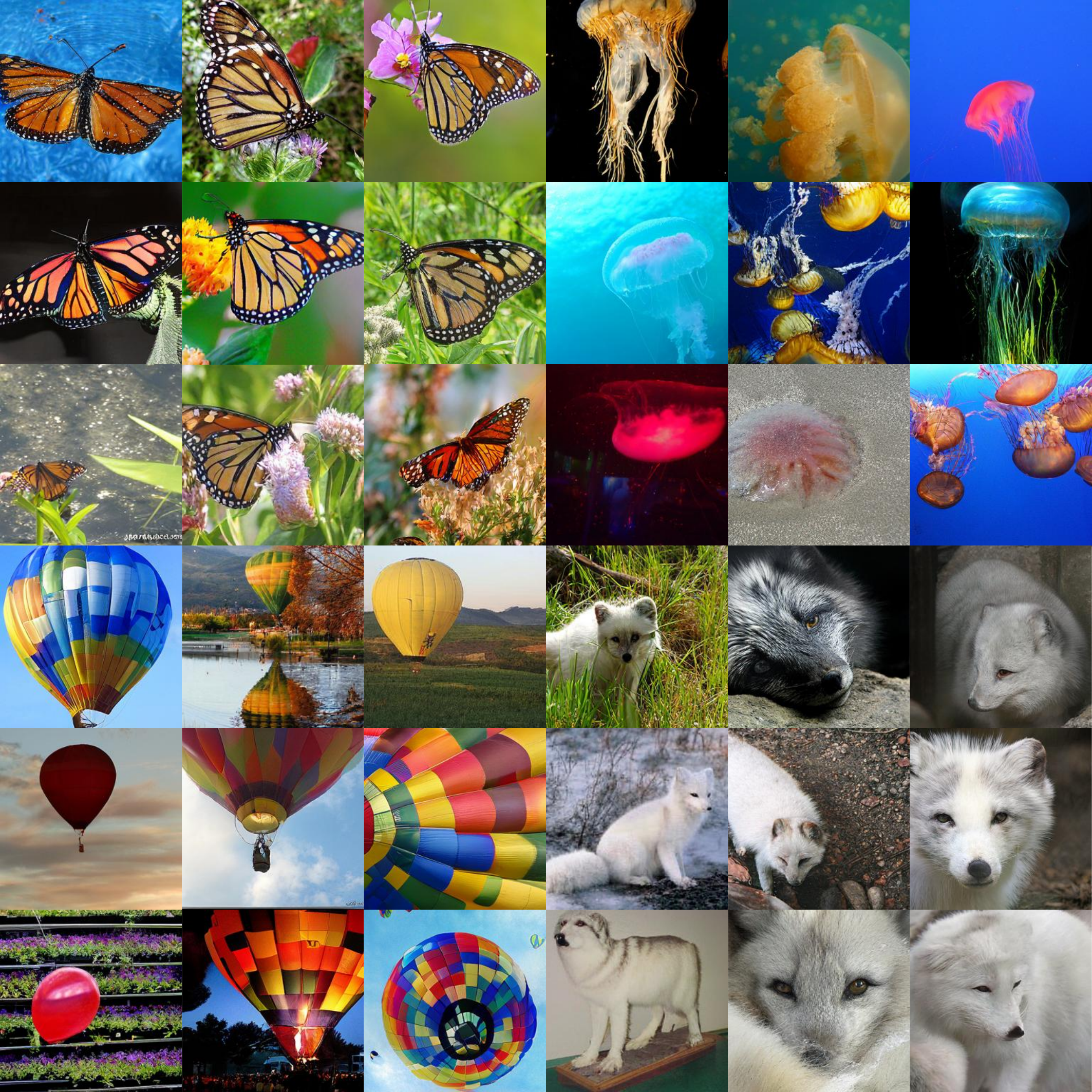}
    \caption{\small{\textbf{More sampled generation results of Infinity-CC + \LSQ (2B).}}
     Classes are 323: monarch butterfly; 107: jellyfish; 417: balloon; 279: arctic fox.}
    \label{fig:more_gen}
\end{figure*}

\myparagraph{Qualitative Results.}
\Cref{fig:more_gen} shows more generation results sampled by Infinity-CC + \LSQ (2B).
We cherry-pick the images and emphasize the quality \textit{and} diversity.

\end{document}